\documentclass[lettersize,journal]{IEEEtran}
\usepackage{amsmath,amsfonts}
\usepackage{algorithmic}
\usepackage{algorithm}
\usepackage{array}
\usepackage[caption=false,font=normalsize,labelfont=sf,textfont=sf]{subfig}
\usepackage{textcomp}
\usepackage{stfloats}
\usepackage{url}
\usepackage{color}
\usepackage{verbatim}
\usepackage{makecell}
\usepackage{graphicx}
\usepackage{colortbl}
\usepackage{multirow}
\usepackage{multirow}
\usepackage{cite}
\usepackage{epstopdf}
\usepackage{gensymb}
\usepackage{multirow}
\usepackage{tabularray}
\begin{document}

\title{Prior-based Objective Inference Mining Potential Uncertainty for Facial Expression Recognition}
\author{Hanwei Liu, Huiling Cai, Qingcheng Lin, Xuefeng Li,~\IEEEmembership{Member,~IEEE,} Hui Xiao
	\thanks{This work was supported by the Shanghai Science and Technology Planning Project (20DZ2205900), Shanghai Municipal Commission of Science and Technology Project (19511132101), Shanghai Municipal Science and Technology Major Project (2021SHZDZX0100), and the Fundamental Research Funds for the Central Universities.}
 
	\thanks{The authors are with College of Electronics and Information Engineering, Tongji University, Shanghai, 201804, China. Xuefeng Li is also with Frontiers Science Center for Intelligent Autonomous Systems, Tongji University. E-mail: liuhw1@tongji.edu.cn, caihuiling@tongji.edu.cn, 1810853@tongji.edu.cn, lixuefeng@tongji.edu.cn, xiaohui@tongji.edu.cn. Corresponding author: Prof. Xuefeng Li, Hui Xiao, Tel: +86 21-69589241, lixuefeng@tongji.edu.cn, xiaohui@tongji.edu.cn}}

\markboth{Journal of \LaTeX\ Class Files,~Vol.~14, No.~8, August~2023}%
{Shell \MakeLowercase{\textit{et al.}}: A Sample Article Using IEEEtran.cls for IEEE Journals}

\IEEEpubid{0000--0000/00\$00.00~\copyright~2023 IEEE}

\maketitle

\begin{abstract}
Annotation ambiguity caused by the inherent subjectivity of visual judgment has always been a major challenge for Facial Expression Recognition (FER) tasks,  particularly for large-scale datasets from in-the-wild scenarios. 
A potential solution is the evaluation of relatively objective emotional distributions to help mitigate the ambiguity of subjective annotations. To this end, this paper proposes a novel Prior-based Objective Inference (POI) network. This network employs prior knowledge to derive a more objective and varied emotional distribution and tackles the issue of subjective annotation ambiguity through dynamic knowledge transfer.
POI comprises two key networks: 
Firstly, the Prior Inference Network (PIN) utilizes the prior knowledge of AUs and emotions to capture intricate motion details. 
To reduce over-reliance on priors and facilitate objective emotional inference, PIN aggregates inferential knowledge from various key facial subregions, encouraging mutual learning.
Secondly, the Target Recognition Network (TRN) integrates subjective emotion annotations and objective inference soft labels provided by the PIN, fostering an understanding of inherent facial expression diversity, thus resolving annotation ambiguity.
Moreover, we introduce an uncertainty estimation module to quantify and balance facial expression confidence. This module enables a flexible approach to dealing with the uncertainties of subjective annotations.
Extensive experiments show that POI exhibits competitive performance on both synthetic noisy datasets and multiple real-world datasets.
All codes and training logs will be publicly available at https://github.com/liuhw01/POI.
\end{abstract}


\section{Introduction}
\IEEEPARstart{F}{acial} expressions, as the most prominent emotionally explicit human feature, can effectively convey emotions and intentions \cite{Ref1,Ref2,Ref3}. Facial expression recognition (FER) technology has a lot of application value in medical diagnosis \cite{Ref4}, human-computer interaction (HCI) \cite{Ref5}, and other application fields. In recent years, FER has made marked progress with improvements in deep learning algorithms, and the emergence of diverse facial expression datasets has accelerated this process, from laboratory datasets such as CK+ \cite{Ref6} and JAFFE \cite{Ref7} to large-scale real-world datasets such as FERPlus \cite{Ref8}, AffectNet \cite{Ref9}, and RAF-DB \cite{Ref10}. 

However, unlike laboratory datasets with fewer subjects and professional actor performances, which lead to high label confidence, annotation ambiguity has been a key challenge for large-scale real-world datasets \cite{Ref69, Ref71,Ref70}. 
First, most real-world datasets use crowd-sourced voting to determine emotional categories, and the subjectivity of observer visual judgement \cite{Ref11,Ref12} leads to inconsistencies in label categories. As shown in Fig. 1(c), for facial expression labeled as \textit{surprise}, most observers recognize it as \textit{happiness}. 
Secondly, partial information loss \cite{Ref13, Ref14}, resulting from issues such as uneven picture quality, blurred images, facial obstructions, and pose variations, leads to inadequate expressions.
Lastly, the expresser may express one or more emotions concurrently \cite{Ref15}, yet most FER datasets only have single-label information.
These uncertainties in annotation lead to blurred emotion category boundaries in the latent space, severely impairing the recognition model's ability to capture discriminative feature embeddings. Therefore, some noise-label tolerant algorithms are needed to reduce their interference with FER.

\begin{figure}[!t]
	\vspace{-0.2cm}
	\centering
	\includegraphics[width=3.1in]{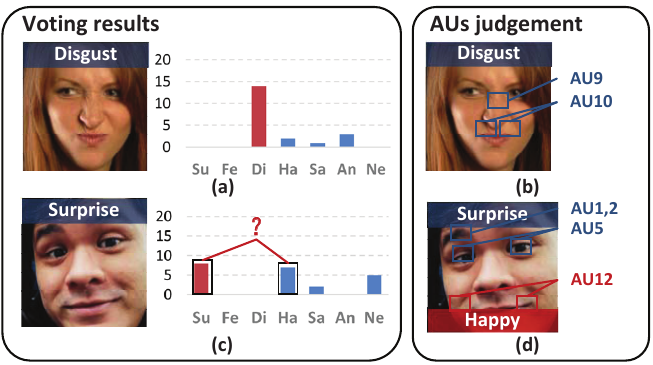}
	\caption{Subjective voting results and AUs judgment results for emotion categories. (a) and (c) show the voting results of 20 volunteers on facial emotion categories, demonstrating the uncertainty of observers' subjective judgment. (b) and (d) show the annotation results of emotion categories based on prior knowledge. Su: Surprise, Fe: Fear, Di: Disgust, Ha: Happy, Sa: Sadness, An: Anger, and Ne: Neutral.}
	\label{Fig_1}
\end{figure}

\IEEEpubidadjcol
Recent studies on annotation ambiguity have primarily focused on strategies such as relabelling \cite{Ref16, Ref17} or label distribution learning \cite{Ref18, Ref19, Ref20, Ref21}. The former substitutes evaluated noisy labels with more precise ones, while the latter leverages latent diversified soft labels to describe samples, based on feature learning.
Unfortunately, these methods often disregard the substantial contribution that objective prior knowledge can offer in mitigating subjective annotation bias.
Numerous studies \cite{Ref22, Ref23, Ref24} have established a notable correlation between facial action units (AUs) and emotions. As illustrated in Fig. 1(b)(d), \textit{disgust} is typically associated with AU9 (nose wrinkle) and AU10 (upper lip raise), and \textit{surprise} is accompanied by AU1 (inner brow raise), AU2 (outer brow raise), and AU5 (upper lip raise) according to prior knowledge. This prior knowledge serves as robust auxiliary information for expression recognition.

Therefore, to address annotation ambiguity, this paper proposes a novel \textbf{P}rior-based \textbf{O}bjective \textbf{I}nference (\textbf{POI}) network. The fundamental concept of POI is to utilize a prior inference network that infers a relatively objective emotion distribution based on AU and emotion prior knowledge. The target recognition network simultaneously learns subjective labels and objective inference distributions, effectively resolving annotation ambiguity.

Specifically, the POI primarily comprises two key networks: the Prior Inference Network (PIN) and the Target Recognition Network (TRN). Firstly, the PIN utilizes prior knowledge of AU and emotions to mine the motion attributes of facial subregions. 
In this network, an Objective Inference Module is designed that aggregates the inferred knowledge of multiple facial subregions for mutual learning, thereby diminishing the over-reliance on AU priors and facilitating the acquisition of relatively objective emotional soft labels.
The TRN learns annotation labels and objective soft labels provided by the PIN, thereby learning the diversity of facial emotions to address annotation ambiguity. In this network, an Uncertainty Estimation Module is designed to quantify and weigh the confidence facial expressions, thereby flexibly assigning objective soft labels to facial expressions to cope with potential confusion in emotion categories.
During the inference phase, only the TRN is utilized, ensuring computational cost is independent of the PIN.

Moreover, there are several methods based on AU and emotions \cite{Ref25,Ref34,Ref35,Ref26,Ref27,Ref28,Ref29,Ref36,Ref37} aim to enhance latent facial representations. These approaches either rely on multi-task joint annotations of AU and emotions  \cite{Ref25,Ref34,Ref35}, which is often infeasible due to the labor-intensive process of AU annotation; or heavily depend on accurate AU prior knowledge from controlled lab datasets \cite{Ref26,Ref27,Ref28,Ref29,Ref36,Ref37}, which can be challenging to attain for real-world datasets of varying quality. For example, as illustrated in Fig.\ref{Fig_1}(d), the facial image is labeled as \textit{surprise}, but the lower half of the face exhibits AU12, which has a strong prior correlation with \textit{happiness}. These methods also face the persistent issue of annotation ambiguity. Different from these methods, the proposed POI network does not require joint annotations of AU and emotions. With its objective inference module and uncertainty estimation module, it can effectively mitigate interference from uncertain prior supervision and resolve annotation confusion.

As a result, our proposed POI effectively resolves annotation ambiguity and delivers competitive results on both popular real-world datasets and synthetic noisy data.
The main contributions are as follows:
\begin{itemize}
	\item This work provides insight to address the confusion caused by subjective annotations by leveraging prior knowledge for objective inference.
        \item This work proposes a novel prior-based objective inference network named POI to address annotation confusion. This includes the development of a prior inference network that discerns the objective distribution of emotions, complemented by a target recognition network. The latter integrates subjective annotations with objective inference, guided by uncertainty estimation.
	\item Extensive quantitative and qualitative experiments validate the effectiveness of POI on popular real-world benchmarks and synthetic noise datasets. 
\end{itemize}

\section{related work}
In this section, we review related research about annotation ambiguity for FER, action units for FER and knowledge distillation.
\subsection{Annotation ambiguity for FER}
Traditional FER methods focus on enhancing deep learning algorithms \cite{Ref13,Ref30,Ref31,zhang2021joint} to uncover latent features with single emotion annotation. However, in real-world scenarios, expressors often display multiple emotional cues \cite{Ref15}, and the visual annotations by observers usually carry subjective biases \cite{Ref11,Ref12}, which complicates the FER task. Recently, increasing research are addressing annotation uncertainty from single annotation data by either re-labeling \cite{Ref16,Ref17} or label distribution learning \cite{Ref18,Ref19,Ref20,Ref21,Ref32,Ref71}. SCN \cite{Ref16} employs a weighting and re-labeling mechanism for each training sample to suppress label uncertainty. DMUE \cite{Ref18} resolves annotation ambiguity by mining latent distributions and uncertainty estimates through multiple emotion category branches. FENN \cite{Ref19} utilizes a multivariate normal distribution to mitigate heteroscedastic uncertainty caused by inter-class label noise. Different from these methods, ours emphasizes the contribution of objective prior knowledge to annotation confusion, resolving annotation ambiguity by inferring objective emotional distributions through prior supervision.
\subsection{Action units for FER}
Facial action units (AUs) \cite{Ref33} are used to describe subtle facial muscle movements and closely related to emotions. Some studies  focus on enhancing latent facial representations based on AU and expression prior knowledge, which can be divided into methods based on external datasets \cite{Ref25,Ref34,Ref35} and prior analysis \cite{Ref26,Ref27,Ref28,Ref29,Ref36,Ref37} to obtain relevant knowledge. The former trained AU detectors based on an external dataset containing AU annotations, and TMSAU-Net \cite{Ref25} and Zhang et al. \cite{Ref34} trained an AU detector via an in-lab spontaneous facial expression 3D video database (BP4D). 
The latter impose weak supervision based on prior knowledge to learn latent embeddings. AUE-CRL \cite{Ref27} mines useful AU representations based on prior knowledge of AU-emotions and mutually exclusive coexistence relations between AUs. FER-IK \cite{Ref29} encodes a Bayesian network based on the probabilistic dependencies of AU-emotions. MER-auGCN \cite{Ref36} and AU-GCN \cite{Ref26} recognize microexpressions based on prior knowledge. However, these methods are usually highly dependent on accurate AU supervision under controlled lab datasets, limiting their generalizability in complex and varying real-world datasets. Notably, they still encounter annotation confusion. Unlike these methods, ours efficiently addresses uncertain prior supervision based on an objective inference module and resolves annotation confusion.

\begin{figure*}[!t]
	\vspace{-0.2cm}
	\centering
	\includegraphics[width=6.5in]{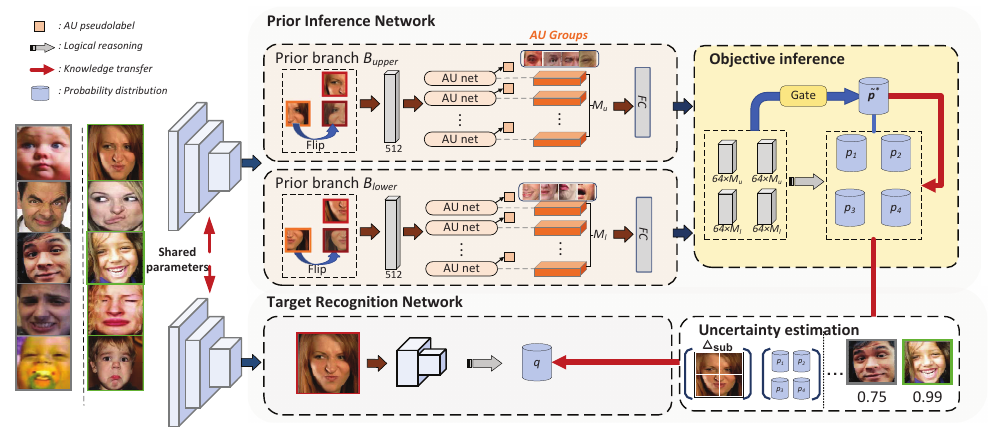}
	\caption{Structure of proposed method. POI consists of a shared feature extractor, a Prior Inference Network, and a Target Recognition Network, which are designed to learn potentially diverse emotions and thereby address annotation ambiguity. ${p_{1}}$, ${{p}_{2}}$, …, ${p_{4}}$ denote emotion prediction distribution of discriminant regions; ${\widetilde{p}^{*}}$ denotes intermediate prediction distribution; and ${q}$ denotes prediction distribution of target network.}
	\label{Fig_2}
\end{figure*}

\subsection{Knowledge distillation}
Knowledge distillation (KD) transfers the knowledge of a well-trained complex model to a smaller network to improve the model's performance \cite{Ref38}.  In addition to the cross-entropy loss based on real labels, distillation also aims to introduce the extra supervision of the teacher model into the student model. Previous distillation methods typically use offline learning strategies, requiring at least two stages of training. The more recently proposed deep mutual learning \cite{Ref39} overcomes this limitation by conducting online distillation in one-phase training between two peer student models. Moreover, distributed learning with Co-distillation has demonstrated the potential benefits of online distillation.  In \cite{Ref40}, each student model learned from the average of the predictions generated by a group of students and achieved a better teacher model effect. Batra et al. \cite{Ref41} proposed cooperative learning, which jointly trains multiple models in different fields. However,  learning average predictions can easily lead to homogenization. Unlike these methods, ours integrates semantic information from each branch based on the Objective Inference module and assigns importance weights via Uncertainty Estimation, reducing the occurrence of homogenization issues during the KD process.

\section{methods}
To address annotation ambiguity, we propose a novel Prior-based Objective Inference (POI) network to facilitate the inference of objective emotion knowledge. In this section, we provide an overview of POI and then introduce its two network branches.

\subsection{Overview of POI}
Fig. \ref{Fig_2} provides an overview of the proposed POI method. It consists of a shared feature extractor $G$, a Prior Inference Network (PIN), and a Target Recognition Network (TRN). 
For a given facial expression image $I$, it first goes through the shared feature extractor $G$ to obtain the latent features $g$. These latent features $g$ are then fed into both the PIN and TRN.  
The PIN is composed of two prior branches and an Objective Inference module (OIM). The prior branches learn the latent movements of facial subregions based on prior knowledge of Action Units (AU) and emotions. Then, the OIM combines their intermediate predictions to facilitate mutual learning and reduce over-reliance on priors, resulting in relatively objective soft labels representing the distribution of emotions. 
The TRN integrates the annotated labels and the objective soft labels inferred by PIN to resolve annotation confusion. Particularly, an Uncertainty Estimation module (UEM) is designed to quantify the confidence of the facial expressions, enabling flexible weight assignment to objective soft labels. During the inference phase, the PIN is removed and only the TRN is used for logical inference, eliminating the need for additional computational cost.

\subsection{Shared feature extractor}
To obtain low-level facial features, for a given ${i}$-th face image $I_i$, it is input into a shared feature extractor $G$ to obtain a latent feature representation $g_i$ with size $R \times L\times L$, that is

\begin{equation}
	\begin{aligned}
	{g_i}=G(I_i)
	\end{aligned}
\end{equation}

\subsection{Prior inference network}
Emotional expression can be regarded as a joint action of facial AUs, reflecting relatively objective emotional information. To uncover the latent distribution of facial expressions, the Prior Inference Network (PIN), which consists of two prior branches and an Objective Inference Module (OIM), infers objective emotional distribution soft labels based on AU and emotion prior knowledge.

\subsubsection{Prior branches}

The prior branches are designed to learn the latent motion attributes of multiple key facial subregions based on weak supervision from AU-emotion, thereby assisting the objective inference module in inferring diverse emotion distributions.

First, to force the PIN to better locate facial muscle movements, we follow the method of Ref \cite{Ref13,Ref72}. 
As shown in Fig.3, the active muscle movements in emotional expression are usually located near the eyes and mouth of the face\cite{Ref42, Ref43, Ref44}. 
Consequently, we segment the facial region into multiple key subregions that have a strong correlation with emotions. 
More specifically, 
we crop the local areas of size ${R\times L_{sub}\times L_{sub}}$ in the upper left, upper right, lower left, and lower right corners of the low-level feature $g$ as key facial local subregions ${\Delta_{sub}}$, where ${g_{\Delta_{sub}} \in { g_{right\_eye}, g_{left\_eye}, g_{right\_mouth}, g_{left\_mouth}}}$. 
Subsequently, due to the facial symmetry, we flip the right part of the feature face, so that the left and right parts of the face have the similar facial structure, that is: 
\begin{equation}
	\begin{aligned}
		{g_{right\_ eye}}^{*} &  = {\rm{Flip}}\left( g_{right\_ eye} \right)\\
		{g_{right\_ mouth}}^{*} &  = {\rm{Flip}}\left( g_{right\_ mouth} \right)
	\end{aligned}
\end{equation}

Next, due to the similar facial structures of $g_{left\_eye}$ and ${g_{right\_eye}}^{*}$, as well as $g_{left\_mouth}$ and ${g_{right\_mouth}}^{*}$, they are separately inputted into the structurally similar but non-shared parameter prior branches ${B_{upper}}$ and ${B_{lower}}$. Specifically, ${B_{upper}}$ receives $g_{left\_eye}$ and ${g_{right\_eye}}^{*}$, while ${B_{lower}}$ takes $g_{left\_mouth}$ and ${g_{right\_mouth}}^{*}$, as shown in Fig.2, to individually learn the motion features of each facial subregion. Each prior branch initially includes a $3\times3$ convolutional layer and a global average pooling (GAP) layer, which outputs 512-dimensional shallow features $F_{\Delta_{sub}}$.

\textit{AU-Expression prior supervision:} The process of emotional expression is often accompanied by specific muscle movements \cite{Ref24}. Therefore, based on the prior knowledge of AU-expression correlation, AU pseudolabels with a high correlation can be set for different emotion categories, guiding the network to distinguish subtle facial muscle movements. 
As summarized in prior knowledge \cite{Ref24,Ref29}, the high correlation between AU and expression is shown in Table I, where the expression-dependent marginal probability of AUs is larger than 70$\%$. These high-correlation AU pseudolabels can provide relatively objective prior supervision, assisting in emotion inference.

\begin{figure}[!t]
	\vspace{-0.2cm}
	\centering
	\includegraphics[width=3in]{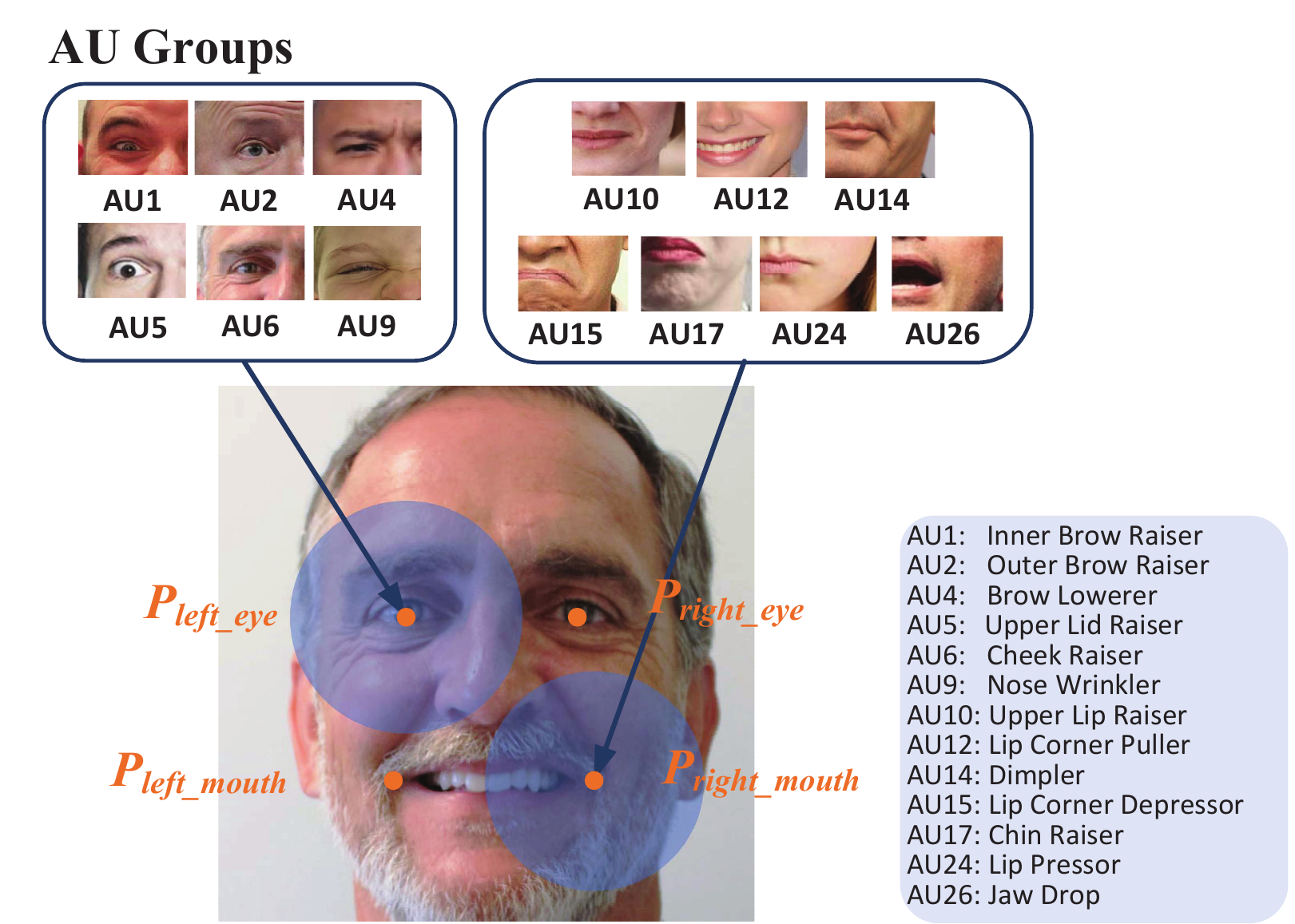}
	\caption{AU groups with high correlation with emotion in key regions of left eye and right side of mouth, which show active muscle movements near eyes and mouth.}
	\label{Fig_3}
\end{figure}

\begin{table}
	\caption{AU-EXPRESSION CORRELATION PRIOR KNOWLEDGE BASED ON \cite{Ref24,Ref29} \label{tab:table1}}
	\vspace{-0.2cm}
	\centering
	\begin{tabular}{c|cc} 
		\hline
		\multirow{2}{*}{Expression} & \multicolumn{2}{c}{Facial regions}  \\ 
		\cline{2-3}
		& Eye           & Mouth               \\ 
		\hline
		Happy                       & AU6           & AU12, AU26          \\
		Fear                        & AU1, AU4, AU5 & AU26                \\
		Anger                       & AU4           & AU24                \\
		Surprise                    & AU1, AU2, AU5 & AU26                \\
		Sadness                     & AU1, AU4      & AU15, AU17          \\
		Disgust                     & AU9           & AU10, AU17          \\
		Contempt                    & /             & AU14                \\
		\hline
	\end{tabular}
\end{table}

In our design, the shallow features $F_{\Delta_{sub}}$ of each facial subregion are connected to $M_{sub}$ AU nets, where $M_{sub} \in \{ M_{u},M_{l}\}$, to decompose the shallow facial features into latent muscle movement features $F_{au}$. Each AU network is designed to learn a specific AU category. Specifically, according to Table I, we set $M_{u}=6$ for $B_{upper}$, learning AU1, AU2, AU4, AU5, AU6, and AU9 respectively, and we set $M_{l}=7$ for $B_{lower}$, learning AU10, AU12, AU14, AU15, AU17, AU24, and AU26 respectively, as shown in Fig.3.

The $m$-th AU net is represented as follows: $m \in \left\{ 1,2,\ldots,M_{sub} \right\}$, for the shallow feature ${I^{n}}_{\Delta_{sub}}$ with $\mathcal{R}^{512}$ of the $n$-th subregion, it is decomposed into AU latent features $I^{n,m}$ with $\mathcal{R}^{128}$ after being connected to the fully connected (FC) layer. 
To improve the differential attention of AUs with different emotional correlations, each $I^{n,m}$ is connected to an FC layer and Sigmoid activation function to get the importance weight:
\begin{equation}
	\begin{aligned}
	&{I_{*}}^{n,m} = Sigmoid\left( {W_{f}I^{n,m}} \right)*I^{n,m}
	\end{aligned}
\end{equation}
where ${I_{*}}^{n,m}$ is the weighted AU feature, $W_{f}$ is the weight of the FC layer, and all ${I_{*}}^{n,m}$ together constitute the AU group. Then, each ${I_{*}}^{n,m}$ is dimensionally reduced to $\mathcal{R}^{64}$ by connecting to an FC layer to acquire different AU's latent representations.

Especially, to supervise the learning of AU, a one-dimensional FC layer and sigmoid function are used for each ${I_{*}}^{n,m}$ to estimate the occurrence probability ${\hat{p}}_{i,n,m}$. 
That is, if the $i$-th face image belongs to the $c$-th expression category, then we deploy the AU loss for each ${\hat{p}}_{i,n,m}$:
\begin{equation}
\begin{split} 	
\mathcal{L}^{AU} = - \frac{1}{K*N_{sub}*M_{sub}}\sum\limits_{i = 1}^{K}\sum\limits_{n = 1}^{N_{sub}}{\sum\limits_{m = 1}^{M_{sub}} \hat{y}_{i,n,m}}log{\hat{p}}_{i,n,m} \\
+ ( 1 - {\hat{y}}_{i,n,m} )log( 1 - log{\hat{p}}_{i,n,m} ) 
\end{split} 	
\end{equation}
where $K$ is the training batch size, $N_{sub}$ is the number of facial subregions, and $\hat{y}_{i,n,m}$ is the pseudolabel of the $m$-th AU of the $n$-th subregion of the $i$-th image after smoothing. The pseudolabel is set according to Table I, AU pseudolabel ${\hat{y}}_{i,n,m} = 1 - \varepsilon$ when the AU categories represented by $m$-th AU net appear in the $c$-th emotional category, and $0+\varepsilon$ otherwise. $\varepsilon$ is used to suppress the influence of false AU pseudolabels under uncertain emotional annotations. By default, $\varepsilon=0.1$ is set in this paper. 

\subsubsection{Objective Inference module}

To obtain potential emotional soft labels, the OIM aggregates the latent motion across multiple subregions, guiding their mutual learning to achieve reasoning soft labels.

Specifically, we define the annotation label of the $i$-th training sample $x_i$ as $y_i$, where $y_i \in \left\{ 1,2,\ldots,C \right\}$, and $C$ is the total number of emotion categories. First, the AU feature $I_{AU}$ of the subregion is input in parallel to the $N_{sub}$ FC layer output logic $t_{i,n} = \left( t_{i,n}^{1},t_{i,n}^{2},\ldots,t_{i,n}^{C} \right)$, $n \in \left\{ {1,2,\ldots,N_{sub}} \right\}$. Then, a Softmax layer is used to obtain the probability distribution of emotional prediction $p_{(i,n)}$ of the $n$-th key subregion in the $i$-th image. 

First, cross-entropy loss (CE) is employed to learn the annotation labels:
\begin{equation}
	\begin{aligned}
		&\mathcal{L}_{CE}^{au} = - \frac{1}{K*N_{sub}}{\sum\limits_{i = 1}^{K}{\sum\limits_{n = 1}^{N_{sub}}y_{i}}}logp_{i,n}
	\end{aligned}
\end{equation}

Next, to obtain relatively objective emotional reasoning soft labels, OIM aggregates the latent emotions mapped by muscle movements across all subregions. This forms an intermediate prediction for objective reasoning and guides mutual learning between subregions. This can also be viewed as a form of knowledge distillation based on objective prior knowledge.
The intermediate prediction reduces reliance on uncertain annotation labels, while the mutual learning amongst subregions minimizes dependence on uncertain priors.

Specifically, for the prediction logic $t_{(i,n)}$ of the $n$-th subregion of the $i$-th image, first apply the temperature $T$ to obtain a soft probability distribution ${\overset{\sim}{p}}_{i,n}$:
\begin{equation}
	\begin{aligned}
		&{\overset{\sim}{p}}_{i,n} = \frac{exp\left( t_{i,n}/T \right)}{\sum_{c = 1}^{C}{exp\left( t_{i,n}^{c}/T \right)}}
	\end{aligned}
\end{equation}
an increased temperature $T$ results in a smoother probability distribution. The default value of $T$ in this study is set as 3. 

To effectively aggregate knowledge from diverse subregions and utilize it as an intermediate prediction steering mutual learning among these subregions, we introduce a gate module. This module calculates the weight of each subregion by using their high-level semantic information, thereby enhancing the learning diversity and addressing the homogeneity issue often seen in traditional averaging each prediction. 
Specifically, the gate module concatenates the set of AU features $I_{AU}$ for all $N_{sub}$ subregions, after which the FC and softmax layers are connected to output the subregion weights of the $i$-th image $w_{i,n} = \left( w_{i,1},w_{i,2},\ldots,w_{i,N_{sub}} \right)$. Then, the intermediate prediction distribution ${\overset{\sim}{p}}_{i}^{*}$ for the i-th image is:
\begin{equation}
	\begin{aligned}
		&{\overset{\sim}{p}}_{i}^{*} = {\sum\limits_{n = 1}^{N}{w_{i,n}*{\overset{\sim}{p}}_{i,n}}}
	\end{aligned}
\end{equation}

Subsequently, based on the intermediate prediction, Kullback-Leibler (KL) divergence loss is deployed for each subregion prediction. This guides mutual learning between subregions and results in a relatively objective emotional reasoning soft label  $\overset{\sim}{p}_{i}^{*}$, $n \in \left\{ {1,2,\ldots,N_{sub}} \right\}$, which can be expressed as:
\begin{equation}
   	\begin{aligned}
   		&\mathcal{L}_{KL}^{au} = - \frac{1}{K*N_{sub}}{\sum\limits_{i = 1}^{K}{\sum\limits_{n = 1}^{N_{sub}}{{\overset{\sim}{p}}_{i}^{*}log\frac{{\overset{\sim}{p}}_{i}^{*}}{{\overset{\sim}{p}}_{i,n}}}}}
   	\end{aligned}
\end{equation}                                                                  

\subsection{Target recognition network}

\begin{figure}[!t]
	\vspace{-0.2cm}
	\centering
	\includegraphics[width=3in]{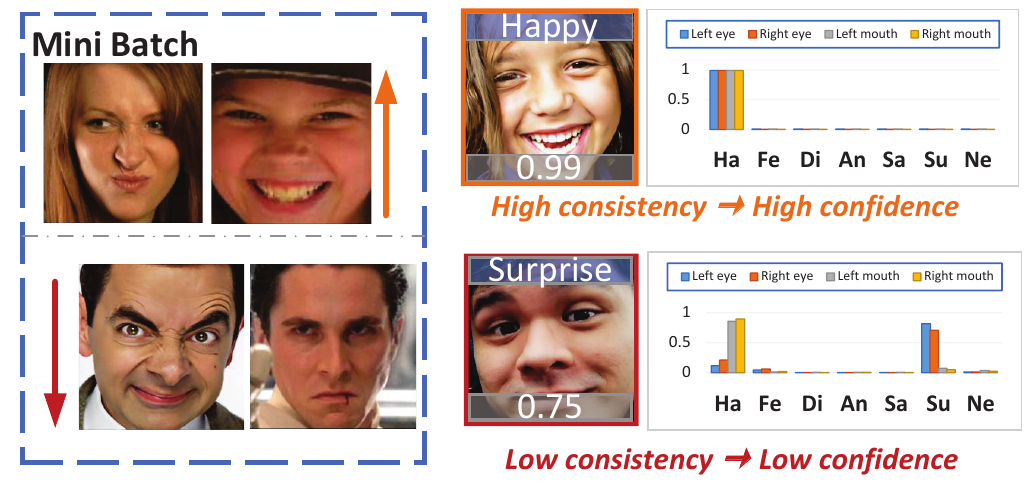}
	\caption{Expression consistency estimation for all facial subregions $\Delta_{sub}$. High (Low) consistency corresponds to high (low) confidence.}
	\label{Fig_4}
\end{figure}

The Target Recognition Network (TRN) learns both subjective annotation labels and objective reasoning soft labels, promoting the understanding of diverse emotions and mitigating annotation ambiguity. Firstly, TRN connects to a feature extractor $G$, which shares parameters with PIN, to yield 512-dimensional emotional features. Next, the FC and softmax layers produce a predictive probability distribution $q_i$ for emotion categories.

Subjective Annotation Labels: Cross-entropy loss is employed to learn subjective annotation labels, which may include certain ambiguous annotations: 
\begin{equation}
	\begin{aligned}
		&\mathcal{L}_{CE}^{tar} = - \frac{1}{K}{\sum\limits_{i = 1}^{K}{y_{i}logq_{i}}}
	\end{aligned}
\end{equation} 

Objective Reasoning Knowledge: To enhance the effectiveness of objective reasoning knowledge transfer, we introduce an Uncertainty Estimation Module (UEM) to measure the potential confidence in facial expressions. Subsequently, based on the uncertainty estimation, TRN flexibly assigns weights to the objective reasoning soft labels of PIN, resolving annotation ambiguity.

\textit{Uncertainty Estimation Module.} 
Intuitively, for different facial regions, when the predicted results tend to be consistent, the results are more credible, while large differences represent unreliable results, as shown in Fig. 4. Therefore, the UEM appraises the potential confidence in facial expressions by assessing the consistency of predictions across $N_{sub}$ subregions in TRN. 
That is, inspired by the uncertainty estimation of Bayesian deep networks \cite{Ref45}, we measure the uncertainty $w_i^{au}$ of the $i$-th image based on epistemic uncertainty:
\begin{equation}
	\begin{aligned}
		&w_{i}^{au} = 1 - \left\lbrack \frac{1}{N_{sub}}{\sum\limits_{n = 1}^{N_{sub}}{( \overset{\sim}{p}_{n})^{2} -}}\left( \frac{1}{N_{sub}}{\sum\limits_{n = 1}^{N_{sub}}{\overset{\sim}{p}}_{n}} \right)^{2} \right\rbrack
	\end{aligned}
\end{equation} 
where $w_i^{au}$ judges the label confidence by measuring the consistency of the subregion predictions, assigning high confidence to face images with high consistency and low confidence to face images with low consistency. 
The visualization results of the uncertainty estimation are demonstrated in Visualization of Uncertainty IV-D.

For facial expressions with low confidence, the model is compelled to learn diverse emotional expressions by assigning a higher weight to objective reasoning soft labels, thereby resolving expression ambiguity. Conversely, for facial expressions with high confidence, the objective reasoning soft labels are assigned a lower weight, enabling the model to focus on consistent emotional categories, that is: 
\begin{equation}
	\begin{aligned}
		\mathcal{L}_{KL}^{tar} & = - \frac{1}{K*N}{\sum\limits_{i = 1}^{K}{\sum\limits_{n = 1}^{N}\left. {\beta_i\left( \overset{\sim}{p} \right.}_{i,n}log\frac{{\overset{\sim}{p}}_{i,n}}{q_{i}} \right)}}\\
		 \beta_i & = (exp(1-{w_{i}^{au}}))-1
	\end{aligned}
\end{equation} 

By doing so, the target recognition network can flexibly learn subjectively annotated emotional labels and objectively inferred soft labels, thus resolving the interference from annotation ambiguity.

\subsection{Overall loss function}
The overall objective of POI is:
\begin{equation}
\mathcal{L}_{total}=\lambda_1\mathcal{L}^{AU}+\lambda_2\mathcal{L}^{au}_{CE}+\lambda_3\mathcal{L}^{tar}_{CE}+T^2(\mathcal{L}^{au}_{KL}+\mathcal{L}^{tar}_{KL})
\end{equation} 
where $\lambda_1$, $\lambda_2$, and $\lambda_3$ are the hyperparameters. In the inference phase, the prior inference network is removed to improve the model's flexibility and reduce the consumption of computational resources.

\section{experiments}
In this section, we describe the datasets used and the implementation details. Then, we conduct qualitative and quantitative analyses to demonstrate the effectiveness of POI in tolerating label noise. Subsequently, we compare POI with existing methods on public datasets. Finally, ablation experiments are performed to demonstrate the effectiveness of each module in POI.

\begin{figure}[!t]
	\centering
	\includegraphics[width=3.1in]{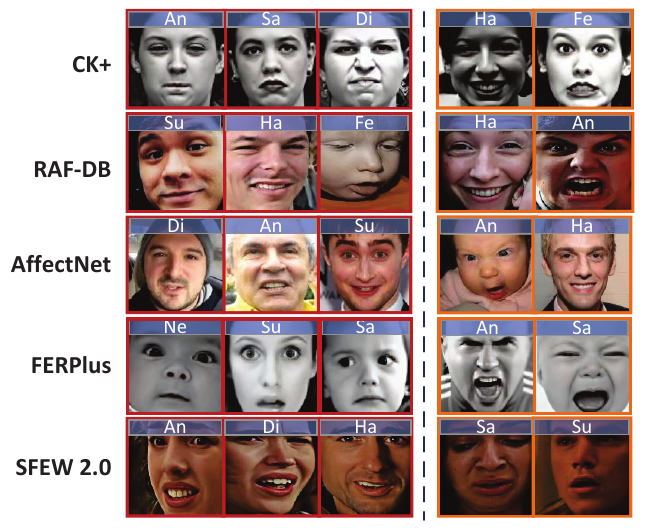}
	\caption{Partial sample of the dataset. The red box indicates that the sample with emotional labelling has certain confusion, and the orange box indicates that the label of the sample has labelling confidence.}
	\label{Fig_5}
\end{figure}

\subsection{Datasets}
In the experiment, we utilized in-the-lab dataset CK+ \cite{Ref6}, as well as in-the-wild datasets including RAF-DB \cite{Ref10}, AffectNet \cite{Ref9}, FERPlus \cite{Ref8}, and SFEW 2.0 \cite{Ref46}, to showcase the effectiveness of POI, particularly its feasibility under real-world scenarios. Fig.5 presents some instances from the datasets.
\textbf{CK+} \cite{Ref6} includes 327 labelled laboratory-controlled facial expression sequences. For each sequence, we use the last three frames as the target emotion and select the first frame as the neutral expression. Finally, 1,236 facial images are used, and 8:2 samples are randomly selected as training and testing sets.
\textbf{RAF-DB} \cite{Ref10} is a large-scale dataset with 30,000 face images and is marked by approximately 40 annotators. In the experiments, we only use images with six basic emotions plus neutral emotions, including 12,271 training images and 3,068 test images.
\textbf{AffectNet} \cite{Ref9} is currently the largest facial emotion dataset, with more than 40 W images. In these experiments, images with six basic emotions plus neutral and contempt emotions are used for testing, which includes 280,000 training images and 4,000 test images.
\textbf{FERPlus} \cite{Ref8} contains approximately 30,000 greyscale images with a size of 40×40 and eight emotion categories with contempt added. We choose 28,709 images as the training set and the public test set with 3,589 images to evaluate the recognition performance.
\textbf{SFEW 2.0} \cite{Ref46} is created by selecting static frames from the AFEW database based on keyframes and includes seven emotion labels, 958 training images, and 436 testing images.

\subsection{Implementation details}
To make fair comparisons with previous studies, by default, we use ResNet-18 as the backbone network pre-trained on VGGFace2 \cite{Ref47}. The auxiliary inference network and target network share the parameters Conv1 \~{} Conv4, and face images are detected and aligned by RetinaFace \cite{Ref48}, and then resized to  224$\times$224 pixels. We use random flip and translation as well as random changes in brightness, contrast, and saturation for data enhancement. In particular, for the CK+ and FERPlus datasets with greyscale images, we do not use data enhancement methods with random changes in brightness, contrast, and saturation. According to the ablation study, the hyperparameters are set by default as $\lambda_1=0.5$, $\lambda_2=0.5$, and $\lambda_3=1.0$. The minibatch size is set to 256 with a momentum of 0.9 and weight decay of 0.0001. The learning rate begins at 0.1 and drops by 10 after 20 epochs. We trained the model for a total of 60 epochs. Stochastic gradient descent (SGD) was used as the optimization algorithm. The number of POI parameters is 60.59 M, and the number of FLOPs is 7.92 G.

Approximately 0.5, 4.5, and 1.0 hours are required to train RAF-DB, AffectNet, and FERPlus, respectively. For the CK+ and SFEW datasets, we pre-train the proposed POI on the RAF-DB dataset and then fine-tune it on CK+ and SFEW for approximately 30 minutes. During the inference phase, the proposed POI achieves a recognition time of 0.94 ms for a single facial image.

\subsection{Annotation ambiguity evaluation}
To verify the effectiveness of our POI in dealing with annotation confusion, we perform quantitative evaluations on the synthetic label noise of RAF-DB, FERPlus, and AffectNet. 
Specifically, we randomly select 10$\%$, 20$\%$, and 30$\%$ of the training data to flip to other emotional categories and compare them with the existing methods that resolve annotation ambiguity, such as SCN \cite{Ref16}, DMUE \cite{Ref18}, LRN \cite{Ref51}, and RUL \cite{Ref71}. For a fair comparison, we utilize the same ResNet-18 used in the benchmark method as the backbone. The results are presented in Table \ref{sys}. Each experiment is conducted six times to determine the average accuracy.

\begin{table}[!t]
	\caption{evaluation of mean precision on synthetic datasets of RAF-DB, FERPlus, and AffectNet datasets ($\%$). $^{*}$ using ResNet-18 as Backbone. \label{sys}}
	\vspace{-0.2cm}
	\centering
	\begin{tabular}{ccccc} 
		\hline\hline
		Method                 & Noisy
		(\%) & RAF-DB                            & FERPlus                           & AffectNet                          \\ 
		\hline\hline
		Baseline               & 10           & 80.52                             & 83.56                             & 57.20                              \\
		SCN \cite{Ref16}               & 10           & 82.18                             & 84.28                             & 58.85                              \\
		DMUE \cite{Ref18}              & 10           & 83.19                             & 83.86                             & 61.21                              \\
		LRN \cite{Ref51}               & 10           & 85.45                             & 84.14                             & 59.49  
                \\ 
            RUL \cite{Ref71}               & 10           & 86.17                             & 86.93                             & 60.54  \\
		\hline
		{\cellcolor[rgb]{0.902,0.902,0.902}}POI\textsuperscript{*}(Ours) & {\cellcolor[rgb]{0.902,0.902,0.902}}10           & {\cellcolor[rgb]{0.902,0.902,0.902}}\textbf{87.48} & {\cellcolor[rgb]{0.902,0.902,0.902}}\textbf{87.60} & {\cellcolor[rgb]{0.902,0.902,0.902}}\textbf{61.78}  \\ 
		\hline\hline
		Baseline               & 20           & 79.01                             & 82.24                             & 56.52                              \\
		SCN \cite{Ref16}               & 20           & 80.10                             & 83.17                             & 57.25                              \\
		DMUE \cite{Ref18}              & 20           & 81.02                             & 83.81                             & 59.06                              \\
		LRN \cite{Ref51}               & 20           & 83.41                             & 84.01                             & 58.28                              \\ 
            RUL \cite{Ref71}               & 20           & 84.32                             & 85.05                             & 59.01  \\
		\hline
		{\cellcolor[rgb]{0.902,0.902,0.902}}POI\textsuperscript{*}(Ours) & {\cellcolor[rgb]{0.902,0.902,0.902}}20           & {\cellcolor[rgb]{0.902,0.902,0.902}}\textbf{86.08} & {\cellcolor[rgb]{0.902,0.902,0.902}}\textbf{86.31} & {\cellcolor[rgb]{0.902,0.902,0.902}}\textbf{60.40}  \\ 
		\hline\hline
		Baseline               & 30           & 75.12                             & 79.34                             & 55.54                              \\
		SCN \cite{Ref16}               & 30           & 77.46                             & 82.47                             & 55.05                              \\
		DMUE \cite{Ref18}              & 30           & 79.41                             & 82.81                             & 56.88                              \\
		LRN \cite{Ref51}               & 30           & 81.84                             & 83.05                             & 57.09                              \\ 
            RUL \cite{Ref71}               & 30           & 82.06                             & 83.90                             & 56.93  \\
		\hline
		{\cellcolor[rgb]{0.902,0.902,0.902}}POI\textsuperscript{*}(Ours) & {\cellcolor[rgb]{0.902,0.902,0.902}}30           & {\cellcolor[rgb]{0.902,0.902,0.902}}\textbf{85.30} & {\cellcolor[rgb]{0.902,0.902,0.902}}\textbf{85.58} & {\cellcolor[rgb]{0.902,0.902,0.902}}\textbf{59.78}  \\
		\hline\hline
	\end{tabular}
	\end{table}	

\begin{figure}[!t]
	\setlength{\abovecaptionskip}{-0.005cm} 
	\setlength{\belowcaptionskip}{-0.2cm}
	\vspace{-0.2cm}
	\centering
	\includegraphics[width=3.2in]{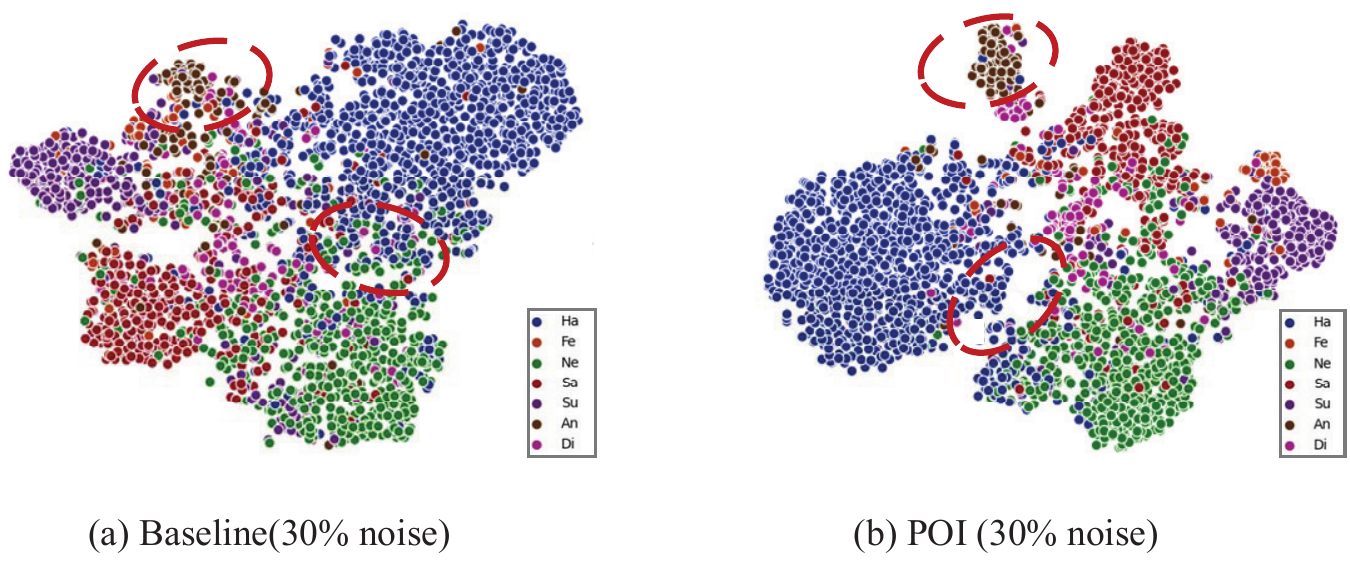}
	\caption{T-SNE visualizations of feature embeddings on the RAF-DB with 30$\%$ noise. Ours produces a clearer category boundary.}
	\label{Fig_7}
\end{figure}

\begin{figure}[!t]
	\setlength{\abovecaptionskip}{-0.005cm} 
	\setlength{\belowcaptionskip}{-0.2cm}
	\vspace{-0.2cm}
	\centering
	\subfloat[]{\includegraphics[width=3.2in]{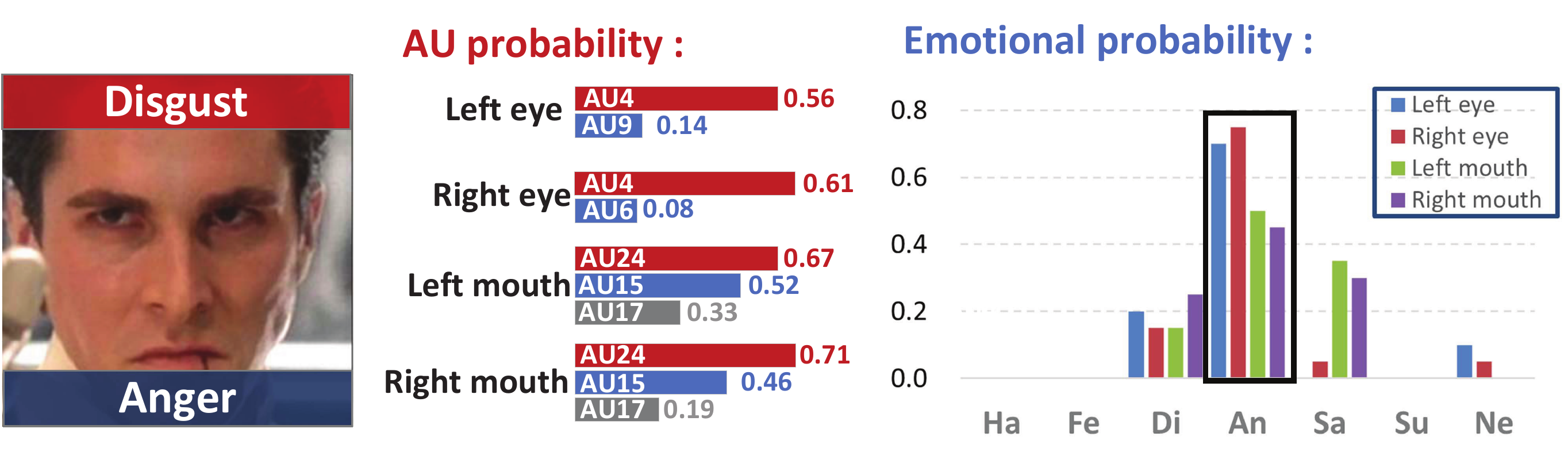}
	}
	\hfil
	\subfloat[]{\includegraphics[width=3.2in]{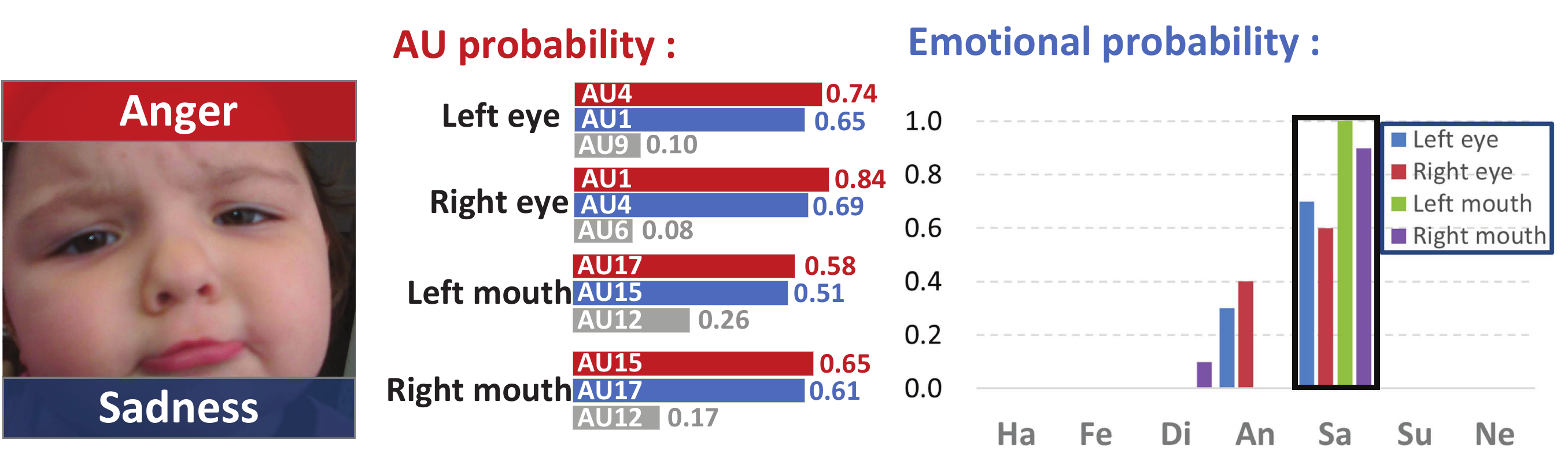}
	}
	\hfil
	\caption{Inference results for mislabelled emotion images in prior inference network. Left: mislabelled image, Middle: Top AU categories predicted by the prior branch, Right: Emotion inference result from the objective inference module. 
  (a) \textit{Anger} (blue) training image mislabelled as \textit{Disgust} (red). (b) \textit{Sadness} (blue) training image mislabelled as \textit{Anger} (red).}
	\label{Fig8}
\end{figure}

As shown in Table \ref{sys}, ours surpasses SCN, DMUE, LRN, and RUL on all datasets. Specifically, ours exceeds RUL by an average of 1.07\%, 1.47\%, and 2.5\% under the conditions of 10\%, 20\%, and 30\% synthetic noise, respectively. Notably, under 30\% annotation noise, which severely hampers the model's ability to extract precise emotional information, our method remains minimally affected,  demonstrating superiority over RUL by 3.24\%, 1.68\%, and 2.85\% on RAF-DB, FERPlus, and AffectNet, respectively. Furthermore, Fig. \ref{Fig_7} showcases the feature distribution of POI under 30\% noise labels, indicating that our method generates more distinct category boundaries. These results validate the effectiveness of our approach in handling noise labels.

Furthermore, Fig. \ref{Fig8} provides a detailed demonstration of POI's inference results under erroneous (confused) annotations, highlighting minimal influence from such annotations on POI. As depicted in Fig. \ref{Fig8}(a), even when an \textit{anger} training image was erroneously tagged as \textit{disgust}, the prior branch in PIN identified faces with pronounced AU4 and AU24 in the facial subregions. Subsequently, the OIM inferred the highly related \textit{anger} category based on these subregion's actions. In Fig. \ref{Fig8}(b), the prior branch identified varying intensities of AU1,4 and AU15,17 in the left and right subregions of the face, enabling the OIM to infer the \textit{sadness} category, which has a high correlation with these AUs,
effectively countering the erroneous emotional category annotation.

\subsection{Objective Inference \& Uncertainty Estimation}
We further validate the proposed Objective Inference Module (OIM) in PIN and the Uncertainty Estimation Module (UEM) in TRN that addresses annotation confusion.

\begin{table}[!t]
	\caption{evaluation of mean precision on synthetic datasets for $w /o$ inpre across RAF-DB, FERPlus, and AffectNet datasets ($\%$).  \label{sys_comp}}
	\vspace{-0.2cm}
	\centering
	\begin{tabular}{ccccc} 
		\hline\hline
		Method                 & Noisy
		(\%) & RAF-DB                            & FERPlus                           & AffectNet                          \\ 
		\hline\hline
            $w /o$ inpre               & 10           & 86.31                             & 86.38                             & 60.45  \\
POI\textsuperscript{*} & 10           & \textbf{87.48} & \textbf{87.60} & \textbf{61.78}  \\ 
		\hline
            $w /o$ inpre               & 20           & 84.03                             & 85.12                             & 58.73  \\
		POI\textsuperscript{*} & 20           & \textbf{86.08} & \textbf{86.31} & \textbf{60.40}  \\ 
		\hline
            $w /o$ inpre               & 30           & 82.49                             & 82.17                             & 57.81  \\
		POI\textsuperscript{*} & 30           & \textbf{85.30} & \textbf{85.58} & \textbf{59.78}  \\
		\hline
	\end{tabular}
	\end{table}	
\textit{\textbf{Objective Inference Module:}} 
As mentioned before, the OIM aggregates intermediate predictions from multiple subregions and guides their mutual learning. Intuitively, there exists another possibility, which is to directly infer the predictions of different facial subregions and use them as emotional soft labels ($w/o$ inpre). 
Table \ref{sys_comp} displays the results of this approach under synthetic noise data. In this setting, we remove ${\overset{\sim}{p}}_{i}^{*}$ and the Gate layer and train OIM solely with $\mathcal{L}_{CE}^{au}$. As a result, ours outperforms $w/o$ inpre by 1.24\%, 1.54\%, and 2.67\% under 10\%, 20\%, and 30\% noise labels, respectively. This validates the effectiveness of the Objective Inference Module. As OIM guides the mutual learning of subregions based on intermediate predictions, it reduces reliance on uncertain prior supervision and confusion in emotional annotations, thereby improving performance.

\begin{table}
\caption{results for test data with varying proportions of different facial expression confidences ($\%$).  \label{sys_pro}}
\vspace{-0.2cm}
\centering
\begin{tabular}{c|ccccccc} 
\hline\hline
\multirow{2}{*}{Dataset} & \multicolumn{7}{c}{Proportion}                                  \\ 
\cline{2-8}
                         & 30\%$\uparrow$           & 50\%$\uparrow$  & 70\%$\uparrow$  & 100\% & 70\%$\downarrow$  & 50\%$\downarrow$  & 30\%$\downarrow$   \\ 
\hline\hline
RAF-DB                   & {\cellcolor[rgb]{0.902,0.902,0.902}}\textbf{99.12} & 96.32 & 93.74 & 89.66 & 85.13 & 80.64 & 74.32  \\
AffectNet                & {\cellcolor[rgb]{0.902,0.902,0.902}}\textbf{74.85} & 70.73 & 68.29 & 62.54 & 59.34 & 56.38 & 53.12  \\
\hline
\end{tabular}
\end{table}

\textit{\textbf{Uncertainty Estimation Module:}} 
To verify the effectiveness of the UEM in assessing the confidence of facial expressions, Table \ref{sys_pro} reports the recognition performance on different proportions of high-confidence ($\uparrow$) and low-confidence ($\downarrow$) test data for the RAF-DB and AffectNet datasets, respectively. The results indicate that recognition performance improves with higher confidence, Specifically, on the RAF-DB and AffectNet datasets, the recognition performance for the 30\% high-confidence subset (30\%$\uparrow$) exceeds that of the 30\% low-confidence subset (30\%$\downarrow$) by 24.8\% and 21.73\%, respectively. 
This clearly demonstrates the efficacy of our proposed approach in measuring expression confidence based on the consistency of facial expressions, particularly in evaluating the confidence of facial expressions without explicit emotion labels.
\begin{figure*}[!t]
	\setlength{\abovecaptionskip}{-0.005cm} 
	\setlength{\belowcaptionskip}{-0.2cm}
	\vspace{-0.2cm}
	\centering
	\subfloat[]{\includegraphics[width=2in]{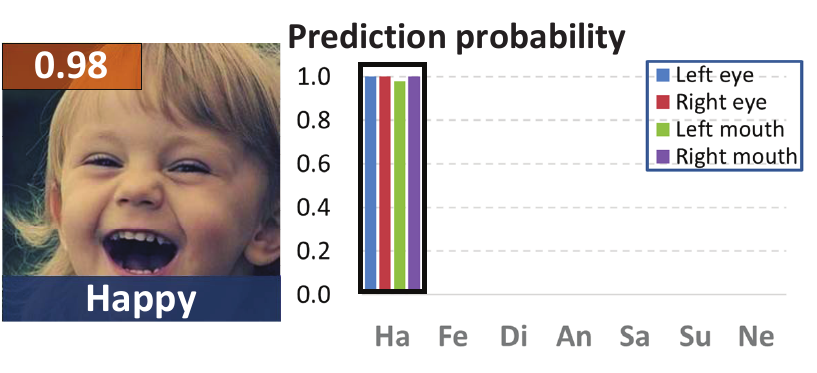}
	}
	\hfil
	\subfloat[]{\includegraphics[width=2in]{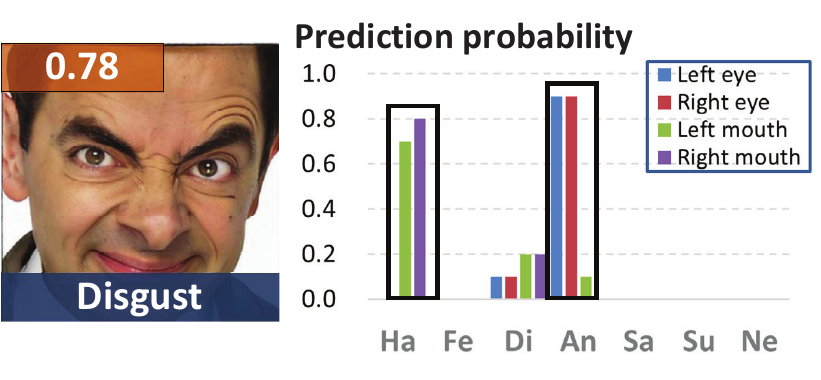}
	}
	\hfil
	\subfloat[]{\includegraphics[width=2in]{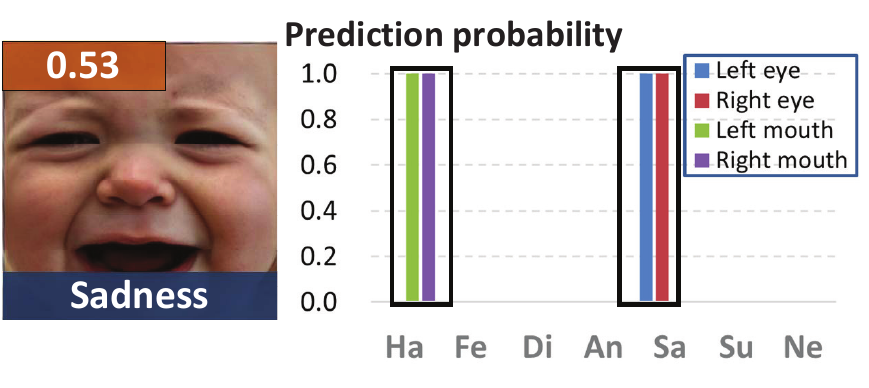}
	}
	\hfil
	\caption{The emotional annotation categories and the potential emotion distribution derived by the PIN at different confidence scores.}
	\label{Fig_10}
\end{figure*}

\begin{figure}[!t]
	\setlength{\abovecaptionskip}{-0.005cm} 
	\setlength{\belowcaptionskip}{-0.2cm}
	\vspace{-0.2cm}
	\centering
	\includegraphics[width=3.5in]{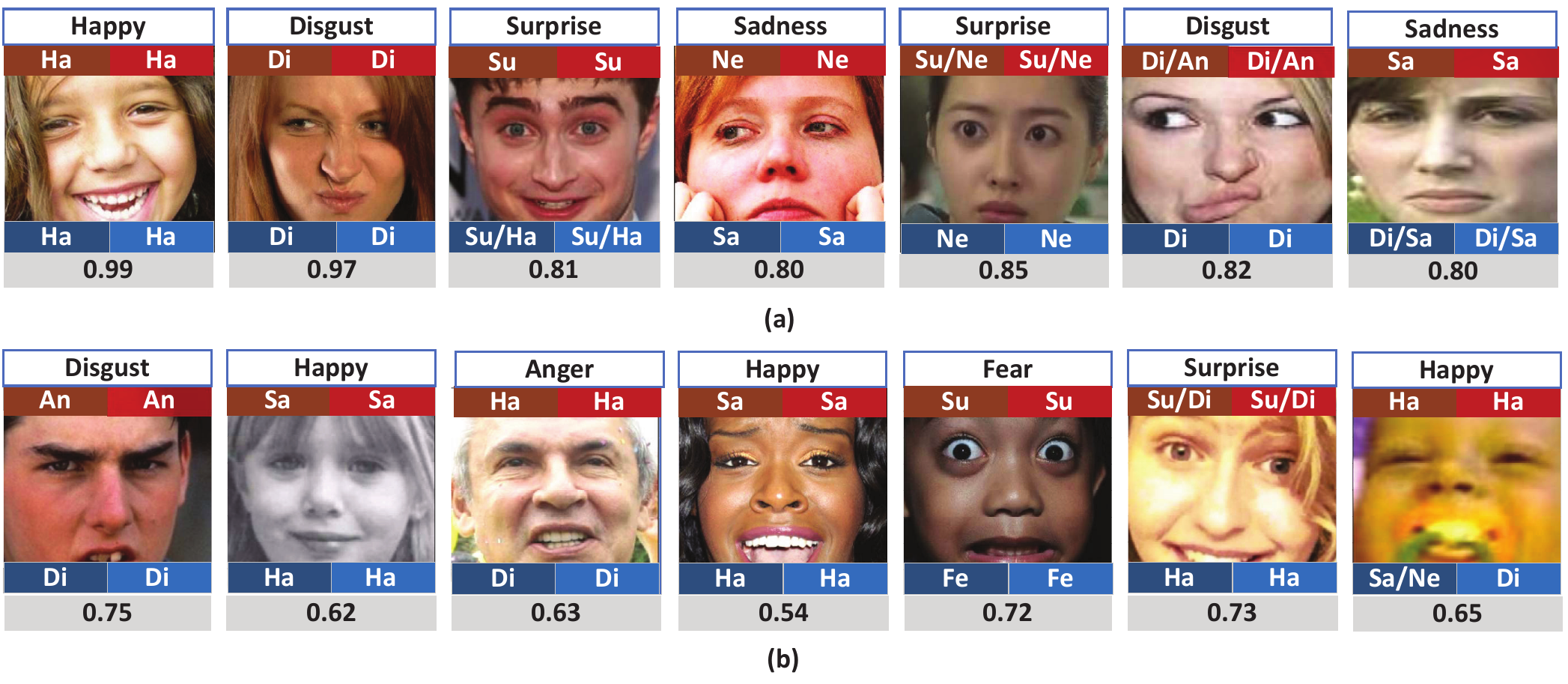}
	\caption{Uncertainty estimation results from UEM alongside the judgment results of potential emotions in various key facial regions by OIM, as applied to a subset of facial images from the RAF-DB and AffectNet datasets.  Grey represents the confidence score, while the blue border is the annotation label. Facial emotion categories at the upper left, upper left, lower left, and lower right represent emotion categories of the left eye, right eye, left mouth, and right mouth based on facial attribute reasoning (probability $\geq0.3$), respectively. (a) Facial images with high emotional confidence, (b) Facial images with low emotional confidence.} 
	\label{Fig_11}
\end{figure}

Furthermore, Fig.\ref{Fig_10} provides a detailed display of the emotional confidence of different images, as well as their inferred emotional soft labels. The confidence of an image is influenced by the consistency of the emotion categories reflected by the muscle movements in various facial regions. Specifically, as shown in Fig.\ref{Fig_10}(a), the facial eye and mouth regions have extremely high consistency because they exhibit AU6, AU12, and AU15, which correspond to \textit{happiness} in prior knowledge. For Fig.\ref{Fig_10}(b), the eyes exhibit AU4 (brow lower) facial attributes corresponding to \textit{anger}, while the mouth with AU12 (lip corner puller) has a strong correlation with \textit{happiness}. Particularly in Fig.\ref{Fig_10}(c), the upper and lower facial parts show distinct emotion categories, resulting in lower confidence. In addition, Fig.\ref{Fig_11} shows the emotional confidence and inferred emotional soft labels for a subset of the RAF-DB and AffectNet datasets. This demonstrates the effectiveness of UEM in quantifying the confidence of facial expressions, as well as the capability of the OIM in inferring potential emotion expressed in different facial subregions.
This further validates their effectiveness in combating emotional confusion.

\begin{figure}[!t]
	\vspace{-0.2cm}
	\setlength{\abovecaptionskip}{-0.005cm} 
	\setlength{\belowcaptionskip}{-0.2cm}
	\centering
	\includegraphics[width=2.5in]{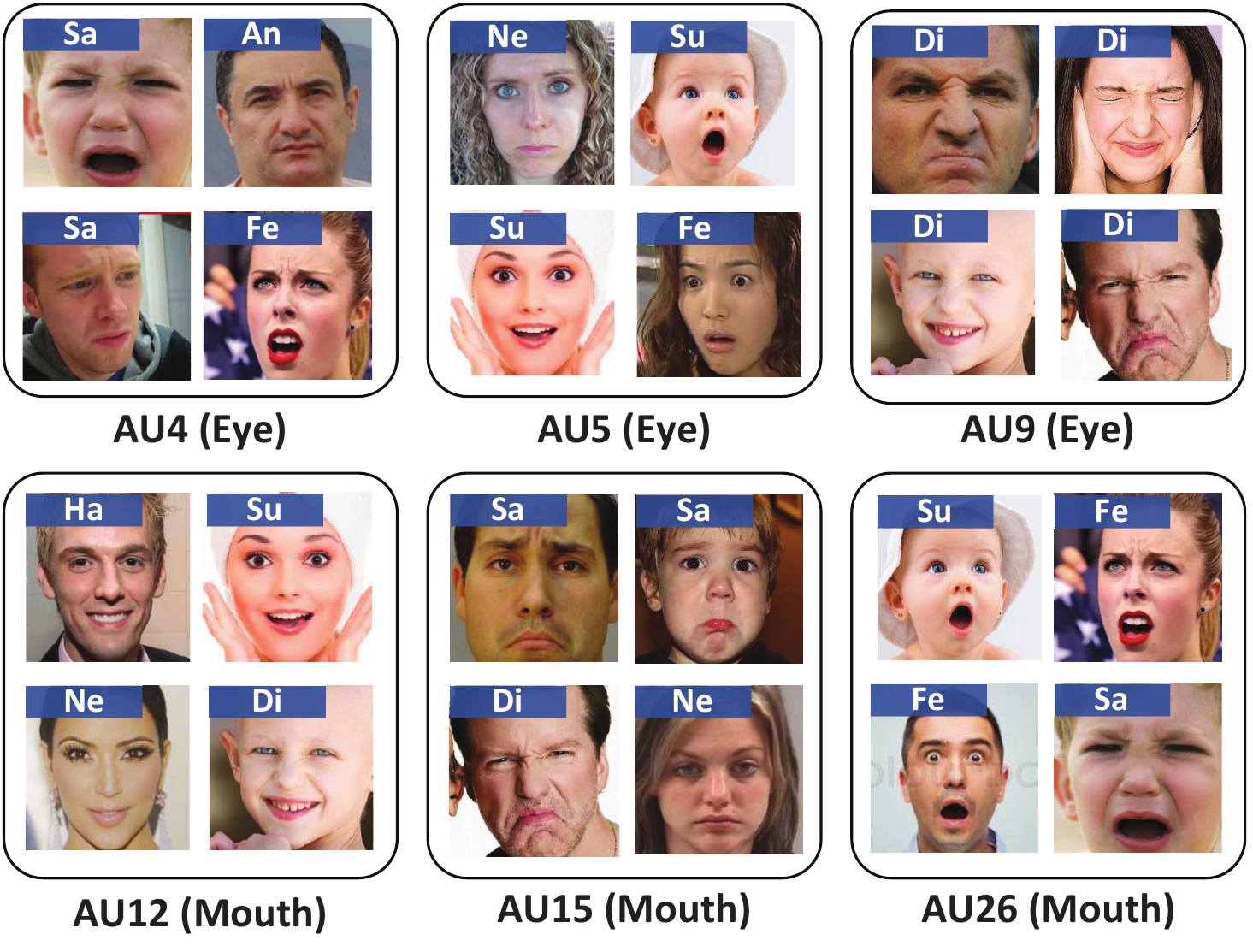}
	\caption{Visualization of AU detection results by the prior branch under RAF-DB and AffectNet datasets. The eye includes results of left and right eyes and the mouth includes results of the left and right sides of the mouth. }
	\label{Fig_9}
\end{figure}
\subsection{Visualization of AU Group}
In this work, the prior branch is utilized to extract facial muscle movement features. Fig.\ref{Fig_9} visualizes the AU detection results of the prior branch for faces with different emotion categories. (1) The prior branch can effectively distinguish muscle movements across different regions, even when those differ from the prior knowledge. For instance, in the AU15 group, faces annotated as \textit{neutral} exhibit the same movement characteristics as AU15 (lip corner depressor). (2) Additionally, Fig.\ref{Fig_9} demonstrates that certain movement attributes are often specific to emotions; for example, when faces express AU9 (nose wrinkle), most are labeled as disgust. (3) Notably, for a given emotion category, different facial subregions may show strong correlations with other emotions. For example, a face labeled \textit{disgust} in the AU12 group displays prominent AU9 in the upper half, which is strongly associated with \textit{disgust}, while the mouth region exhibits distinct AU12 (lip corner puller), which, according to prior knowledge, is strongly associated with \textit{happiness}. These observations reflect the ambiguity of emotional categories and the effectiveness of the prior branch in discriminating muscle movements.

\subsection{Comparison with existing methods}
We compare our method with existing methods on five popular datasets: CK+, RAF-DB, AffectNet, FERPlus, and SFEW 2.0.

\begin{figure*}[!t]
	\setlength{\abovecaptionskip}{-0.005cm} 
	\setlength{\belowcaptionskip}{-0.2cm}
	\vspace{-0.2cm}
	\centering
	\subfloat[RAF-DB]{\includegraphics[width=1.6in]{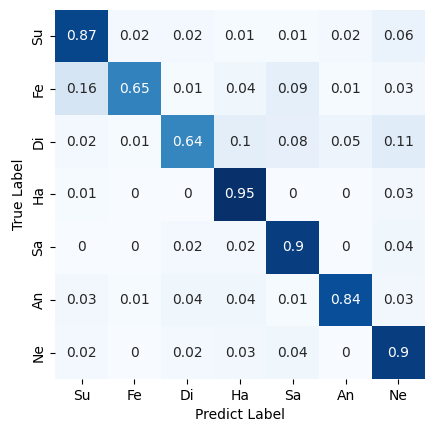}
	}
	\hfil
	\subfloat[AffectNet]{\includegraphics[width=1.6in]{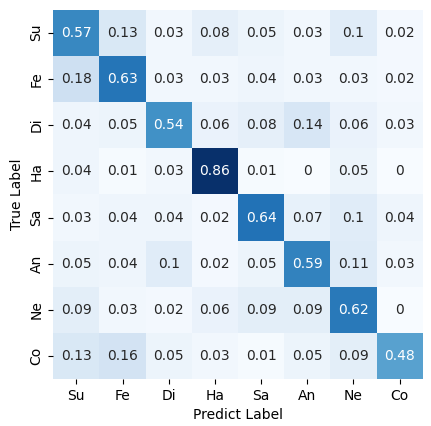}
	}
	\hfil
	\subfloat[FERPlus]{\includegraphics[width=1.6in]{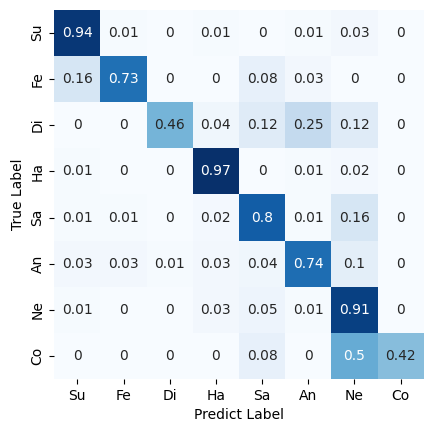}
	}
	\hfil
	\subfloat[SFEW 2.0]{\includegraphics[width=1.6in]{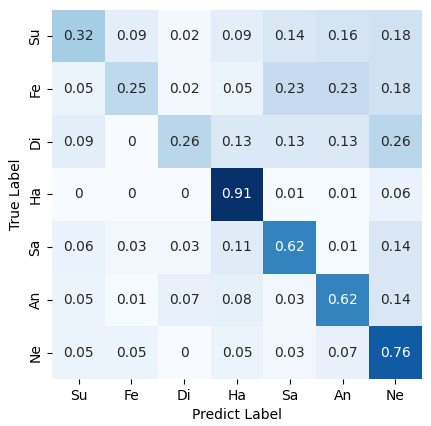}
	}
	\caption{Confusion matrices of (a) RAF-DB, (b) AffectNet, (c) FERPlus, and (d) SFEW 2.0 datasets. Su: Surprise, Fe: Fear, Di: Disgust, Ha: Happiness, Sa: Sadness, An: Anger, Ne: Neutral, Co: Contempt.}
	\label{fig12}
\end{figure*}

\textit{\textbf{Comparison with CK+:}} Table \ref{tab:table5} shows the comparison results between ours and existing methods in the CK+ dataset, which is a dataset collected in the laboratory with high FER quality. Ours achieves the highest recognition accuracy (99.89$\%$), primarily due to the high-quality annotations in the FER dataset, which aid our PIN in obtaining objective reasoning soft labels with higher confidence. In particular, compared to CMCNN \cite{Ref52}, which uses facial landmarks to detect key muscle movements, the proposed method outperforms CMCNN by 1.22$\%$ at the same settings, illustrating POI's excellent ability to uncover underlying emotions.

\begin{table}[!t]
\centering
	\vspace{-0.2cm}
\setlength{\tabcolsep}{2.5mm}{
	\caption{Comparison to existing results on CK+ dataset. $^*$ using ResNet-18 as Backbone \label{tab:table5}}
	\begin{tabular}{cccc} 
		\hline
		Method                 & Backbone   & Acc. (\%) & $\Delta\uparrow$ (\%)
		     \\ 
		\hline
  		Triplet loss \cite{Ref61}          & ResNet-18  & 97.61   & 0.00       \\
		TMSAU-Net \cite{Ref25}              & Manually   & 99.10 & +1.49          \\
		Zhang
		et al. \cite{Ref34}        & DLN        & 98.30    &   +0.69     \\
		CMCNN \cite{Ref52}                 & ResNet-18  & 98.33   &  +0.72      \\
		AFDCNN \cite{Ref59}                & Manually   & 98.43   &  +0.83      \\
		HSIC \cite{Ref60}                  & Inception  & 98.60     & +0.99      \\ 
		\hline
		POI\textsuperscript{*} & ResNet-18     & [99.55]       &  +1.94  \\
		POI                    & ResNet-50     & \textbf{99.89}  & +2.28 \\
		\hline
	\end{tabular}}
	
\end{table}

\begin{table}[!t]
	\caption{Comparison to existing results on RAF-DB dataset.
		\label{tab:table6}}
	\vspace{-0.2cm}
	\centering
     \setlength{\tabcolsep}{2.5mm}{
	\begin{tabular}{cccc} 
		\hline
		Method                 & Backbone    & Acc.
		(\%)  & $\Delta\uparrow$ (\%)   \\ 
		\hline
            TMSAU-Net \cite{Ref25}             & Manually   & 75.80    &0.00       \\
		SCN \cite{Ref16}               & ResNet-18  & 87.03     & +11.23     \\
		Zhang et al. \cite{Ref34}        & DLN         & 85.90    & +10.10       \\
		DMUE \cite{Ref18}              & ResNet-18   & 88.76    &+12.96       \\
		AUE-CRL \cite{Ref27}               & ResNet-101 & 81.00   &+5.20        \\
		SPLDL \cite{Ref54}                 & ResNet-18  & 88.59   &+12.79         \\
		LRN \cite{Ref51}                 & ResNet-18   & 88.91     & +13.11      \\
		WS-FER \cite{Ref53}                & ResNet-50  & 88.89      & +13.09    \\
		PAT-CNN \cite{Ref62}           & ResNet-34  & 86.34       & +10.54   \\
            EAC \cite{Ref68}
            & ResNet-50           & [90.35]       & +14.55    \\
		Face2Exp \cite{Ref64}            & ResNet-50   & 88.54      &+12.74     \\ 
		\hline
		POI\textsuperscript{*} & ResNet-18      & 89.66      & +13.86    \\
		POI                    & ResNet-50      & \textbf{90.51} &+14.71 \\
		\hline
	\end{tabular}}
\end{table}

\begin{table}[!t]
	\caption{Comparison to existing results on AffectNet dataset. \label{tab:table7}}
	\vspace{-0.2cm}
	\centering
      \setlength{\tabcolsep}{2.5mm}{
	\begin{tabular}{cccc} 
		\hline
		Method                 & Backbone  & Acc. (\%)    & $\Delta\uparrow$ (\%)    \\ 
		\hline
            RAN \cite{Ref31}             & ResNet-18   & 59.50  & 0.00 \\
            LRN \cite{Ref51}       & ResNet-18   & 60.83  & +1.33 \\
            SCN \cite{Ref16}               & ResNet-18  & 60.23     & +0.73  \\
            DMUE \cite{Ref18}              & ResNet-18   & [62.84]    & +3.34      \\
		Face2Exp \cite{Ref64}             & ResNet-50 & 60.17     & +0.67     \\
		SPLDL \cite{Ref54}                & ResNet-18  & 60.19    & +0.69     \\ 
		\hline
		POI\textsuperscript{*} & ResNet-18   & 62.54        & +3.04  \\
		POI                    & ResNet-34   & \textbf{63.02}  &+3.52 \\
		\hline
	\end{tabular}}
\end{table}

\textit{\textbf{Comparison with RAF-DB:}} Table \ref{tab:table6} shows the comparison results between the proposed and existing methods in the RAF-DB dataset, and the recognition confusion matrix is shown in Fig.\ref{fig12}(a). POI achieves the highest FER result at 90.51$\%$. Other methods, such as SCN \cite{Ref16}, DMUE \cite{Ref18}, SPLDL \cite{Ref57}, LRN \cite{Ref51}, and EAC \cite{Ref68} are also developed to address annotation ambiguity. Under the same settings, POI outperforms these methods by 2.63\%, 0.9\%, 0.75\%, and 0.16\% respectively.

Moreover, when compared with AU-emotion based methods such as AUE-CRL \cite{Ref27}, TMSAU-Net \cite{Ref25}, and Zhang et al. \cite{Ref34}, our method surpasses their performance by 9.51\%, 14.71\%, and 4.61\% respectively. These results clearly demonstrate the effectiveness of POI for the generalization of uncertain prior supervision and learning underlying emotions.

\textit{\textbf{Comparison with AffectNet:}} 
 Table \ref{tab:table7} shows the comparison results between our method and existing methods in the AffectNet dataset. Fig.\ref{fig12}(b) shows the recognition confusion matrix. Our method achieves the highest recognition result. Specifically, when compared with other methods that also address annotation ambiguity such as LRN, SCN, DMUE, and SPLDL, our method outperforms them by 2.19\%, 2.79\%, 0.18\%, and 2.83\%, respectively. These results underscore the effectiveness of POI.

\textit{\textbf{Comparison with FERPlus:}} Table \ref{tab:table8} shows a comparison between the proposed and existing methods in the FERPlus dataset, which is a set of greyscale facial images with eight emotion categories added with contempt, and the recognition confusion matrix is shown in Fig.\ref{fig12}(c). The proposed method achieves the highest recognition result (90.21$\%$), which is 0.7$\%$ better than the DMUE \cite{Ref18} method using ResNet-50 IBN as the backbone, our method also does not require additional inference.

\textit{\textbf{Comparison with SFEW 2.0:}} Table \ref{tab:table9} shows a comparison between ours and existing methods in the SFEW 2.0 dataset, and the recognition confusion matrix is shown in Fig.\ref{fig12}(d). Ours obtained the best results and outperformed AUE-CRL \cite{Ref27} by 7.06$\%$, which suggests an attentive hybrid architecture based on different face regions. For DMUE \cite{Ref18}, which also solves the problem of annotation ambiguity, our method outperforms it by 2.74$\%$ and 1.52$\%$, respectively. In addition, the results of the proposed method are 5.3$\%$ and 14.08$\%$ higher than WS-FER \cite{Ref56} and CMCNN \cite{Ref52}, respectively, which highlights the FER performance of POI.

\begin{table}[!t]
	\caption{Comparison to existing results on FERPlus dataset \label{tab:table8}}
		\vspace{-0.2cm}
	\centering
      \setlength{\tabcolsep}{2.5mm}{
	\begin{tabular}{cccc} 
		\hline
		Method                         & Backbone      & Acc.
		(\%)  & $\Delta\uparrow$ (\%)   \\ 
		\hline
  		PT \cite{Ref55}                        & ResNet-18     & 86.6   & 0.00 \\
		SCN \cite{Ref16}                        & ResNet-18    & 88.01   & +1.41       \\
		SCN \cite{Ref16}                       & ResNet-50IBN  & 89.35   & +2.75        \\
		DMUE \cite{Ref18}                      & ResNet-18     & 88.64  & +2.04         \\
		DMUE${^\ddag}$ \cite{Ref18} & ResNet-50IBN  & 89.51       & +2.91    \\
		CVT \cite{Ref63}                       & ResNet-18     & 88.81   & +2.21        \\
		LRN \cite{Ref51}                         & ResNet-18     & [89.53] & +2.93          \\ 
		\hline
		POI\textsuperscript{*}         & ResNet-18       & 89.42   & +2.82        \\
		POI                            & ResNet-50       & \textbf{90.21} & +3.61 \\
		\hline
	\end{tabular}}
\end{table}

\begin{table}[!t]
	\caption{Comparison to state-of-the-art results on SFEW 2.0 dataset. ${^\dag}$ and ${^\ddag}$ using ResNet-18 and ResNet-50IBN as Backbone. \label{tab:table9}}
	\vspace{-0.2cm}
	\setlength{\abovecaptionskip}{-0.05cm} 
	\setlength{\belowcaptionskip}{-0.5cm}
	\centering
      \setlength{\tabcolsep}{2.5mm}{
	\begin{tabular}{cccc} 
		\hline
		Method                         & Pretrained
		Dataset & Acc. 
		(\%)  & $\Delta\uparrow$ (\%)   \\ 
		\hline
  		CMCNN \cite{Ref52}                       & MS-Celeb-1
		M       & 45.78   & 0.00        \\
		Zhang
		et al. \cite{Ref34}                 & RAF-DB                & 56.00    & +10.22        \\
		TMSAU-Net \cite{Ref25}                      & BP4D                & 57.40   &+11.62        \\
		AUE-CRL \cite{Ref27}                     & RAF-DB             & 52.80  &+7.02         \\
		DMUE${^\dag}$ \cite{Ref18} & MS-Celeb-1
		M       & 57.12     & +11.34     \\
		DMUE${^\ddag}$ \cite{Ref18} & MS-Celeb-1
		M       & [58.34]    &+12.56       \\
		WS-FER \cite{Ref53}                    & AffectNet          & 54.56      &+8.78     \\
		PAT-CNN \cite{Ref62}                    & ImageNet            & 57.57   &+11.79        \\ 
		\hline
		POI                          & RAF-DB                 & \textbf{59.86} &+14.08  \\
		\hline
	\end{tabular}}
	\end{table}

\subsection{Ablation experiments}
Ablation studies are conducted to verify the contribution of each module in the POI.

\subsubsection{Influence of components}
Table \ref{tab:table3} presents the results of the Prior Branch (PB), Objective Inference Module (OIM), and Uncertainty Estimation Module (UEM) on the RAF-DB and FERPlus datasets. Each component contributes to the results, with the PB significantly improving the performance of the target network on the RAF-DB and FERPlus datasets, achieving an increase of 1.35\% and 2.3\% over the baseline, respectively. This process mines objective attributes based on AU-emotion prior knowledge. The OIM promotes mutual learning of inference knowledge between subregions and reduces the interference of uncertain prior supervision and uncertain emotion annotations. The UEM further weights the confidence of facial expressions, enhancing the transmission of objective reasoning knowledge.

\begin{table}[!t]
	\caption{Evalution of each module in POI on RAF-DB and FERPlus datasets ($\%$), TB stands for Target Backbone. \label{tab:table3}}
	\vspace{-0.2cm}
	\centering
       \setlength{\tabcolsep}{3.0mm}{
	\begin{tabular}{cccc|cc} 
		\hline
		\multirow{2}{*}{TB} & \multicolumn{3}{c|}{Modules} & \multicolumn{2}{c}{Dataset}  \\ 
		\cline{2-6}
		& PB & OIM & UEM                    & RAF-DB & FERPlus             \\ 
		\hline
		\multirow{7}{*}{\checkmark}  & -  & -  & -                            & 86.78  & 86.05               \\
		& \checkmark  & ~  & ~                            & 88.13  & 88.35               \\
		& ~  & \checkmark  & ~                            & 87.64  & 87.06               \\
		& ~  & ~  & \checkmark                            & 87.82  & 87.34               \\
		& \checkmark  & \checkmark  & ~                            & 89.29  & 89.27               \\
		& \checkmark  & ~  & \checkmark                            & 88.82  & 89.07               \\
		& ~  & \checkmark  & \checkmark                            & 88.02  & 88.28               \\ 
		\hline
		\checkmark                   & \checkmark  & \checkmark  & \checkmark                            & 89.66  & 89.42               \\
		\hline
	\end{tabular}}
\end{table}
\subsubsection{Influence of parameters}
\begin{table}[!t]
	\caption{Impacts of region size $L_{sub}$ on RAF-DB dataset.  \label{tab:table4}}
	\vspace{-0.2cm}
	\setlength{\abovecaptionskip}{-0.005cm} 
	\setlength{\belowcaptionskip}{-0.2cm}
	\centering
        \setlength{\tabcolsep}{3.0mm}{
	\begin{tabular}{cccccc} 
		\hline
		$L_{sub}$ & 7     & 8     & 9     & 10    & 11     \\
            \hline
		Acc.       & 89.14 & 89.37 & 89.66 & 89.42 & 89.08  \\ 
		\hline
	\end{tabular}}
\end{table}

\begin{table}[!t]
	\caption{Impacts of temperature $T$ on RAF-DB dataset.  \label{tab:table9}}
	\vspace{-0.2cm}
	\setlength{\abovecaptionskip}{-0.005cm} 
	\setlength{\belowcaptionskip}{-0.2cm}
	\centering
        \setlength{\tabcolsep}{3.0mm}{
	\begin{tabular}{cccccc} 
		\hline
		\textit{T} & 1     & 2     & 3     & 4     & 5      \\ 
		\hline
		Acc.       & 88.79 & 88.86 & 89.66 & 89.32 & 88.81  \\
		\hline
	\end{tabular}}
\end{table}

\begin{figure}[!t]
	\setlength{\abovecaptionskip}{-0.005cm} 
	\setlength{\belowcaptionskip}{-0.2cm}
	\vspace{-0.2cm}
	\centering
	\includegraphics[width=2.7in]{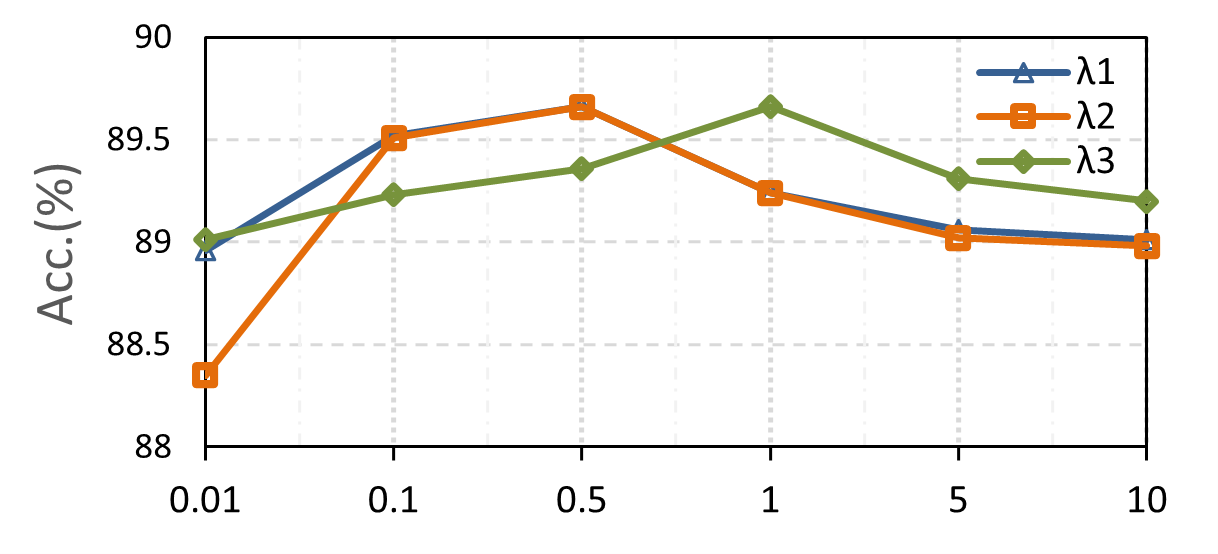}
	\caption{Impacts of weights $\lambda_1$, $\lambda_2$, and $\lambda_3$  on performance of RAF-DB dataset.}
	\label{Fig_6}
\end{figure}

\textit{\textbf{Region size $L_{sub}$:}} 
Table \ref{tab:table4} reports the FER results for different $L_{sub}$ in the PIN. In this setup, we set $L$ to 7, 8, 9, 10, and 11 to evaluate the FER performance. The FER results increase with the rise of $L_{sub}$, achieving the highest result (89.66\%) when $L_{sub}=9$. When $L_{sub} < 9$, we believe that smaller subregions may miss some essential AU information, whereas when $L_{sub} > 9$, we posit that, although the increase in region size enhances the integrity of the key region, larger key regions may lead to inaccurate AU attribute localization.

\textit{\textbf{Weight $\lambda_1$, $\lambda_2$, and $\lambda_3$ :}} 
Fig.\ref{Fig_6} shows the FER results for different values of $\lambda_1$, $\lambda_2$, and $\lambda_3$. In this setup, we first set $\lambda_1=0.5$, $\lambda_2=0.5$, and $\lambda_3=1$, and change the value of a single parameter each time while keeping the remaining values constant. Here, $\lambda_1$ is related to prior knowledge supervision. Constraints that are too small may reduce the ability to distinguish facial muscle movement attributes, while constraints that are too large may decrease generalizability. $\lambda_2$ and $\lambda_3$ are related to the learning ability of the PIN and TRN to annotate label knowledge. Compared to TRN, PIN requires less constraint from annotated labels to infer the objective distribution of emotions.

\textit{\textbf{Temperature $T$:}} 
Table \ref{tab:table9} reports the impact of different values of $T$. $T$ determines the amount of information carried by the soft target during the knowledge transfer process. The higher the $T$, the smoother the output probability distribution and the greater the distribution entropy. As shown in the table, our method shows the highest recognition results when $T=3$. When $T<3$, the distribution quickly becomes steep, and the model is more sensitive to the information carried by the soft target. When $T>3$, the soft target carries too little information, which might reduce the effectiveness of knowledge transfer.

\subsection{Performance of PIN Vs TRN}
\begin{figure}[!t]
	\vspace{-0.4cm}
	\setlength{\abovecaptionskip}{-0.005cm} 
	\setlength{\belowcaptionskip}{-0.2cm}
	\centering
	\includegraphics[width=2.9in]{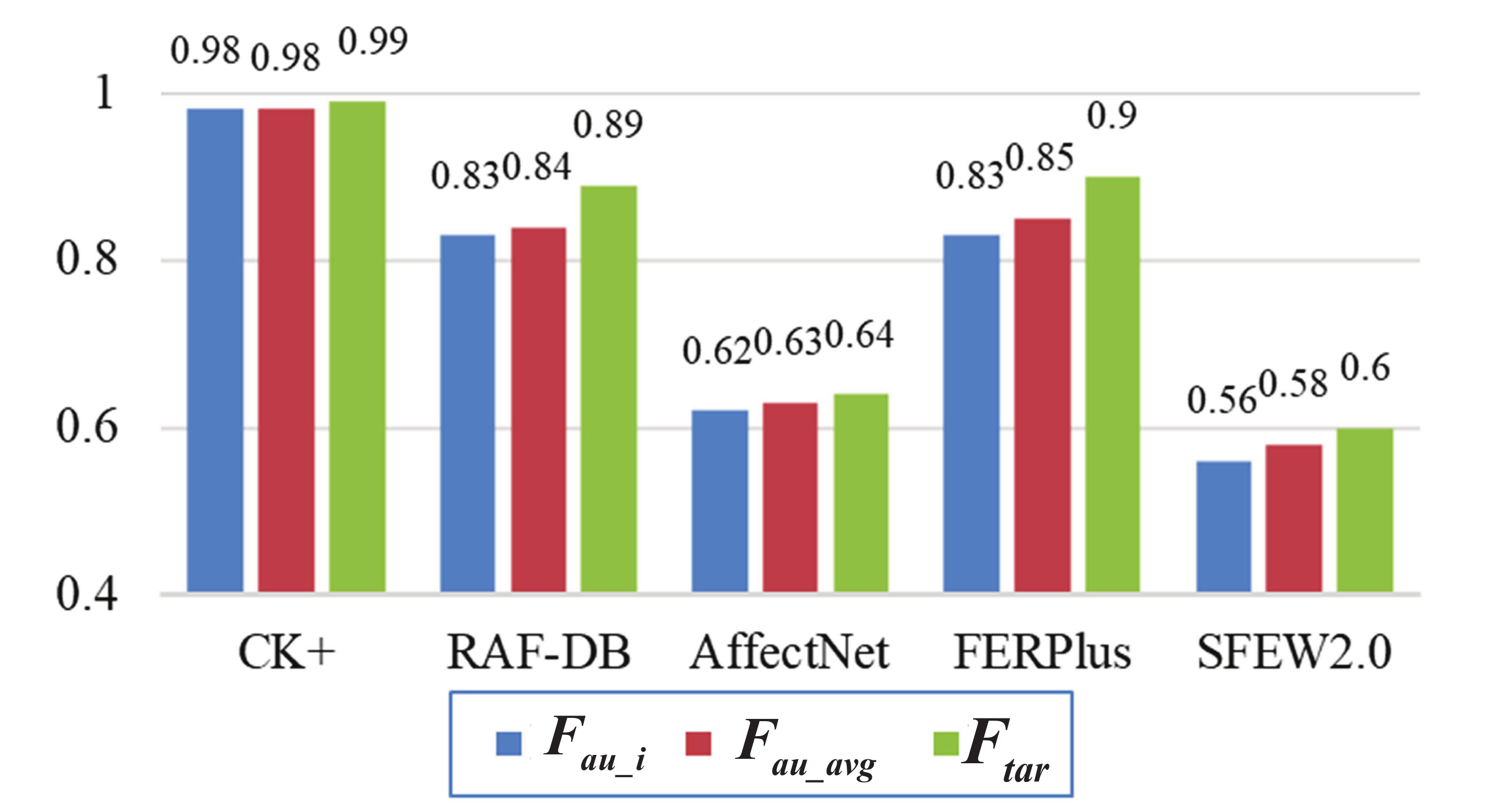}
	\caption{Accuracy with different fusion strategies of the auxiliary inference network and the target network.} 
	\label{Fig_13}
\end{figure}

In this section, we report the performance of the PIN and TRN, as illustrated in Fig.\ref{Fig_13}. Specifically, we present the predictive performance of three different strategies: (1) $F_{au\_i}$: the intermediate prediction of the PIN, ${\overset{\sim}{p}}_{i}^{*}$; (2) $F_{au\_avg}$: the average prediction of facial subregions in the PIN, that is, $( {\sum\limits_{n = 1}^{N_{sub}}{\overset{\sim}{p}}_{n}} /N_{sub})$; (3) $F_{tar}$: the prediction result of the TRN. It can be seen that $F_{tar}$ outperforms $F_{au\_i}$ and $F_{au\_avg}$ across all datasets. This is because $F_{tar}$ learns both annotated labels and objective inference labels, effectively resolving annotation confusion. 
In contrast, $F_{au\_i}$ and $F_{au\_avg}$ focus on inferring a relatively objective emotion distribution based on prior supervision, resulting in lower performance than $F_{tar}$. 
Additionally, the overall performance of $F_{au\_avg}$, derived from intermediate predictions, is superior to that of $F_{au\_i}$, demonstrating the benefits of mutual learning among facial subregions based on intermediate predictions.

\begin{figure}[!t]
	\vspace{-0.4cm}
	\setlength{\abovecaptionskip}{-0.005cm} 
	\setlength{\belowcaptionskip}{-0.2cm}
	\centering
	\includegraphics[width=2.5in]{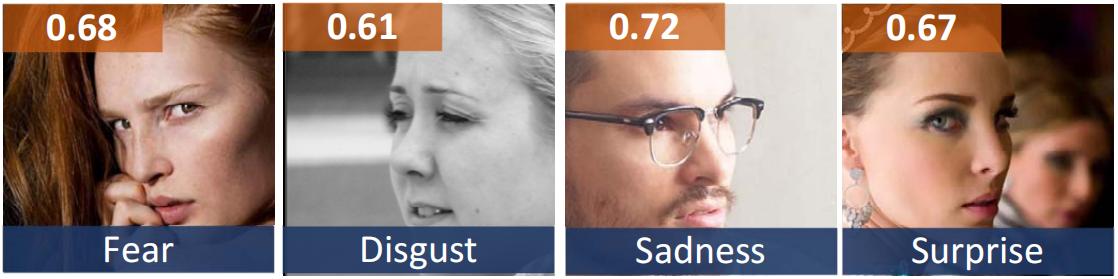}
	\caption{Failed examples. The blue box indicates the label, and the orange represents emotional confidence.} 
	\label{Fig_fail}
\end{figure}

\subsection{Failure examples}
Fig.\ref{Fig_fail} presents some failure examples in POI. For faces with extremely large poses, POI tends to assign them low confidence, even though these faces have a certain emotional confidence. Large poses also poses a significant challenge in FER tasks, due to the considerable loss of critical facial information.
It is worth mentioning that POI is effective in addressing another challenge: facial occlusion,  as illustrated by the last example in the second row of Fig.\ref{Fig_11}. This is because the prior branch efficiently extracts potential features from different key sub-regions of the face.

\begin{figure}[!t]
	\vspace{-0.4cm}
	\setlength{\abovecaptionskip}{-0.005cm} 
	\setlength{\belowcaptionskip}{-0.2cm}
	\centering
	\includegraphics[width=3.1in]{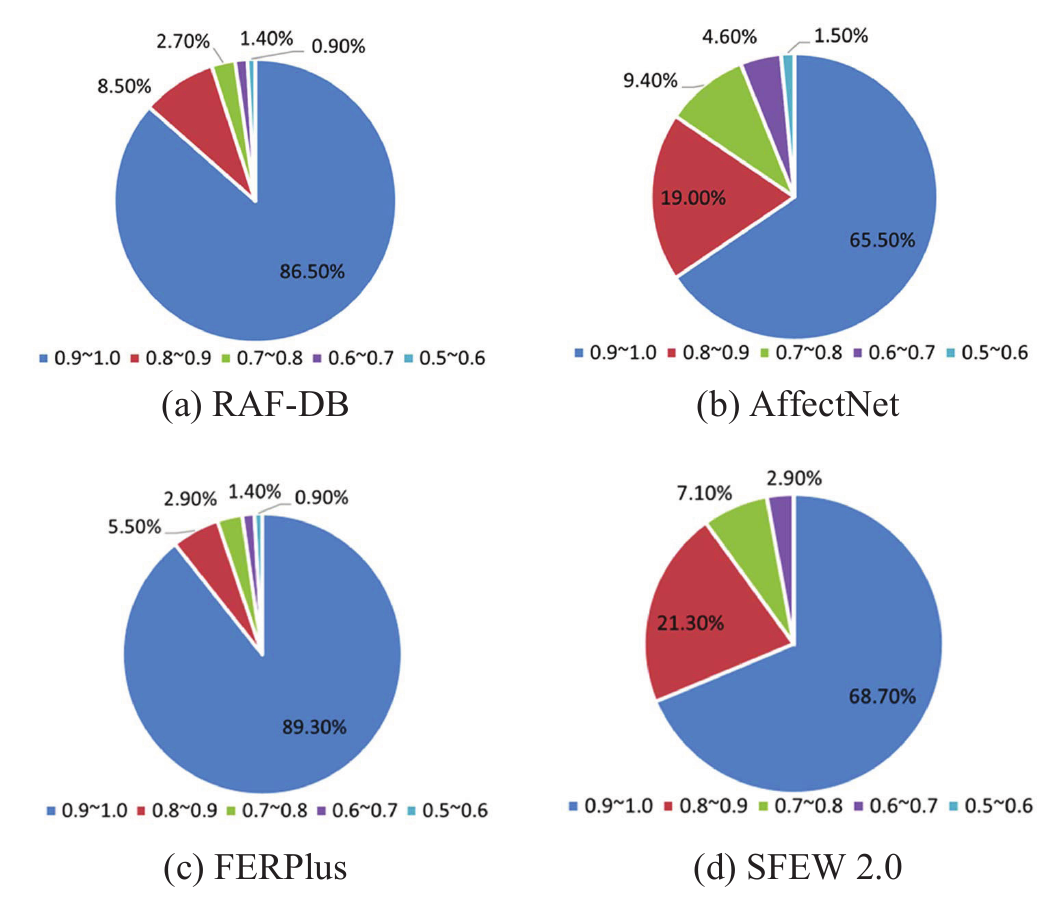}
	\caption{Statistical results of uncertain confidence scores of different FER datasets. }
	\label{Fig14}
\end{figure}

\subsection{Further analysis}
In this section, we further analyze the association between annotation confusion and recognition performance to better address the bottleneck of FER in real-world scenarios. Specifically, we statistically analyze the confidence level of facial expressions in different real-world FER datasets based on uncertainty estimation, as shown in Fig.\ref{Fig14}. The statistics clearly reveal two points: (1) For the RAF-DB and FERPlus datasets, which exhibit excellent recognition performance (see Table \ref{tab:table6} and \ref{tab:table8}), the majority of facial images (86.5\%, 89.3\%) have high emotional confidence. (2) Conversely, for the AffectNet and SFEW 2.0 datasets, which show lower recognition performance (see Table \ref{tab:table7} and \ref{tab:table9}), confused expressions account for a significant proportion, specifically 34.5\% and 31.3\%, respectively, indicating a correlation between the confusion in facial expressions and recognition performance. Furthermore, Fig.\ref{Fig14} also reflects that for real-world FER datasets, a single emotion category annotation may not fully cover the facial emotional categories. The more complex the facial emotional expression, the greater the likelihood of recognition confusion. Notably, compared with single emotion annotation, compound emotion annotation can more effectively describe rich facial emotional information. This aligns with the concept behind our method, that is, addressing the significant impact of label confusion by learning the diversity of facial expressions.

\section{Conclusion}
This work proposes a prior-based objective inference (POI) to solve the annotation ambiguity in FER. POI can extract objective muscle movement attributes of different facial regions and promote emotional knowledge reasoning to learn diverse emotional distributions. Experimental results on synthetic noisy and real-world datasets demonstrated the superiority and effectiveness of the proposed approach, highlighting that reduced ambiguous label weights and mining potential diverse emotions can promote the model's understanding of emotions and thereby improve FER performance. Related research concepts can also provide references for other emotion recognition based on physiological information or explicit behavior and improve generalizability by mining latent emotion annotations.

\bibliographystyle{IEEEtran}
\bibliography{refer}

\begin{thebibliography}{10}
\providecommand{\url}[1]{#1}
\csname url@samestyle\endcsname
\providecommand{\newblock}{\relax}
\providecommand{\bibinfo}[2]{#2}
\providecommand{\BIBentrySTDinterwordspacing}{\spaceskip=0pt\relax}
\providecommand{\BIBentryALTinterwordstretchfactor}{4}
\providecommand{\BIBentryALTinterwordspacing}{\spaceskip=\fontdimen2\font plus
\BIBentryALTinterwordstretchfactor\fontdimen3\font minus \fontdimen4\font\relax}
\providecommand{\BIBforeignlanguage}[2]{{%
\expandafter\ifx\csname l@#1\endcsname\relax
\typeout{** WARNING: IEEEtran.bst: No hyphenation pattern has been}%
\typeout{** loaded for the language `#1'. Using the pattern for}%
\typeout{** the default language instead.}%
\else
\language=\csname l@#1\endcsname
\fi
#2}}
\providecommand{\BIBdecl}{\relax}
\BIBdecl

\bibitem{Ref1}
S.~Li and W.~Deng, ``Deep facial expression recognition: A survey,'' \emph{IEEE transactions on affective computing}, 2020.

\bibitem{Ref2}
J.~Lee, S.~Kim, S.~Kim, and K.~Sohn, ``Multi-modal recurrent attention networks for facial expression recognition,'' \emph{IEEE Transactions on Image Processing}, vol.~29, pp. 6977--6991, 2020.

\bibitem{Ref3}
N.~Perveen, D.~Roy, and K.~M. Chalavadi, ``Facial expression recognition in videos using dynamic kernels,'' \emph{IEEE Transactions on Image Processing}, vol.~29, pp. 8316--8325, 2020.

\bibitem{Ref4}
C.~Bisogni, A.~Castiglione, S.~Hossain, F.~Narducci, and S.~Umer, ``Impact of deep learning approaches on facial expression recognition in healthcare industries,'' \emph{IEEE Transactions on Industrial Informatics}, vol.~18, no.~8, pp. 5619--5627, 2022.

\bibitem{Ref5}
R.~Aylett, C.~Ritter, M.~Y. Lim, F.~Broz, P.~E. McKenna, I.~Keller, and G.~Rajendran, ``An architecture for emotional facial expressions as social signals,'' \emph{IEEE Transactions on Affective Computing}, vol.~12, no.~2, pp. 293--305, 2019.

\bibitem{Ref6}
P.~Lucey, J.~F. Cohn, T.~Kanade, J.~Saragih, Z.~Ambadar, and I.~Matthews, ``The extended cohn-kanade dataset (ck+): A complete dataset for action unit and emotion-specified expression,'' in \emph{2010 ieee computer society conference on computer vision and pattern recognition-workshops}.\hskip 1em plus 0.5em minus 0.4em\relax IEEE, 2010, pp. 94--101.

\bibitem{Ref7}
M.~Lyons, S.~Akamatsu, M.~Kamachi, and J.~Gyoba, ``Coding facial expressions with gabor wavelets,'' in \emph{Proceedings Third IEEE international conference on automatic face and gesture recognition}.\hskip 1em plus 0.5em minus 0.4em\relax IEEE, 1998, pp. 200--205.

\bibitem{Ref8}
E.~Barsoum, C.~Zhang, C.~C. Ferrer, and Z.~Zhang, ``Training deep networks for facial expression recognition with crowd-sourced label distribution,'' in \emph{Proceedings of the 18th ACM International Conference on Multimodal Interaction}, 2016, pp. 279--283.

\bibitem{Ref9}
A.~Mollahosseini, B.~Hasani, and M.~H. Mahoor, ``Affectnet: A database for facial expression, valence, and arousal computing in the wild,'' \emph{IEEE Transactions on Affective Computing}, vol.~10, no.~1, pp. 18--31, 2017.

\bibitem{Ref10}
S.~Li, W.~Deng, and J.~Du, ``Reliable crowdsourcing and deep locality-preserving learning for expression recognition in the wild,'' in \emph{Proceedings of the IEEE conference on computer vision and pattern recognition}, 2017, pp. 2852--2861.

\bibitem{Ref69}
Z.~Wu and J.~Cui, ``La-net: Landmark-aware learning for reliable facial expression recognition under label noise,'' in \emph{Proceedings of the IEEE/CVF International Conference on Computer Vision}, 2023, pp. 20\,698--20\,707.

\bibitem{Ref71}
Y.~Zhang, C.~Wang, and W.~Deng, ``Relative uncertainty learning for facial expression recognition,'' \emph{Advances in Neural Information Processing Systems}, vol.~34, pp. 17\,616--17\,627, 2021.

\bibitem{Ref70}
H.~Liu, H.~Cai, Q.~Lin, X.~Li, and H.~Xiao, ``Learning from more: Combating uncertainty cross-multidomain for facial expression recognition,'' in \emph{Proceedings of the 31st ACM International Conference on Multimedia}, 2023, pp. 5889--5898.

\bibitem{Ref11}
C.~Chen and R.~E. Jack, ``Discovering cultural differences (and similarities) in facial expressions of emotion,'' \emph{Current opinion in psychology}, vol.~17, pp. 61--66, 2017.

\bibitem{Ref12}
R.~E. Jack, C.~Blais, C.~Scheepers, P.~G. Schyns, and R.~Caldara, ``Cultural confusions show that facial expressions are not universal,'' \emph{Current biology}, vol.~19, no.~18, pp. 1543--1548, 2009.

\bibitem{Ref13}
H.~Liu, H.~Cai, Q.~Lin, X.~Li, and H.~Xiao, ``Adaptive multilayer perceptual attention network for facial expression recognition,'' \emph{IEEE Transactions on Circuits and Systems for Video Technology}, vol.~32, no.~9, pp. 6253--6266, 2022.

\bibitem{Ref14}
D.~Poux, B.~Allaert, N.~Ihaddadene, I.~M. Bilasco, C.~Djeraba, and M.~Bennamoun, ``Dynamic facial expression recognition under partial occlusion with optical flow reconstruction,'' \emph{IEEE Transactions on Image Processing}, vol.~31, pp. 446--457, 2021.

\bibitem{Ref15}
S.~Du, Y.~Tao, and A.~M. Martinez, ``Compound facial expressions of emotion,'' \emph{Proceedings of the national academy of sciences}, vol. 111, no.~15, pp. E1454--E1462, 2014.

\bibitem{Ref16}
K.~Wang, X.~Peng, J.~Yang, S.~Lu, and Y.~Qiao, ``Suppressing uncertainties for large-scale facial expression recognition,'' in \emph{Proceedings of the IEEE/CVF conference on computer vision and pattern recognition}, 2020, pp. 6897--6906.

\bibitem{Ref17}
S.~Zhang, Z.~Huang, D.~P. Paudel, and L.~Van~Gool, ``Facial emotion recognition with noisy multi-task annotations,'' in \emph{Proceedings of the IEEE/CVF Winter Conference on Applications of Computer Vision}, 2021, pp. 21--31.

\bibitem{Ref18}
J.~She, Y.~Hu, H.~Shi, J.~Wang, Q.~Shen, and T.~Mei, ``Dive into ambiguity: Latent distribution mining and pairwise uncertainty estimation for facial expression recognition,'' in \emph{Proceedings of the IEEE/CVF Conference on Computer Vision and Pattern Recognition}, 2021, pp. 6248--6257.

\bibitem{Ref19}
Y.~Gu, H.~Yan, X.~Zhang, Y.~Wang, Y.~Ji, and F.~Ren, ``Towards facial expression recognition in the wild via noise-tolerant network,'' \emph{IEEE Transactions on Circuits and Systems for Video Technology}, 2022.

\bibitem{Ref20}
S.~Chen, J.~Wang, Y.~Chen, Z.~Shi, X.~Geng, and Y.~Rui, ``Label distribution learning on auxiliary label space graphs for facial expression recognition,'' in \emph{Proceedings of the IEEE/CVF conference on computer vision and pattern recognition}, 2020, pp. 13\,984--13\,993.

\bibitem{Ref21}
S.~Wang, H.~Shuai, C.~Liu, and Q.~Liu, ``Bias-based soft label learning for facial expression recognition,'' \emph{IEEE Transactions on Affective Computing}, 2022.

\bibitem{Ref22}
F.~Zhang, T.~Zhang, Q.~Mao, and C.~Xu, ``A unified deep model for joint facial expression recognition, face synthesis, and face alignment,'' \emph{IEEE Transactions on Image Processing}, vol.~29, pp. 6574--6589, 2020.

\bibitem{Ref23}
B.~Martinez, M.~F. Valstar, B.~Jiang, and M.~Pantic, ``Automatic analysis of facial actions: A survey,'' \emph{IEEE transactions on affective computing}, vol.~10, no.~3, pp. 325--347, 2017.

\bibitem{Ref24}
S.~Wang, G.~Peng, and Q.~Ji, ``Exploring domain knowledge for facial expression-assisted action unit activation recognition,'' \emph{IEEE Transactions on Affective Computing}, vol.~11, no.~4, pp. 640--652, 2018.

\bibitem{Ref25}
L.~Liang, C.~Lang, Y.~Li, S.~Feng, and J.~Zhao, ``Fine-grained facial expression recognition in the wild,'' \emph{IEEE Transactions on Information Forensics and Security}, vol.~16, pp. 482--494, 2020.

\bibitem{Ref34}
W.~Zhang, Z.~Guo, K.~Chen, L.~Li, Z.~Zhang, Y.~Ding, R.~Wu, T.~Lv, and C.~Fan, ``Prior aided streaming network for multi-task affective analysis,'' in \emph{Proceedings of the IEEE/CVF International Conference on Computer Vision}, 2021, pp. 3539--3549.

\bibitem{Ref35}
V.~Suresh and D.~C. Ong, ``Using positive matching contrastive loss with facial action units to mitigate bias in facial expression recognition,'' in \emph{2022 10th International Conference on Affective Computing and Intelligent Interaction (ACII)}.\hskip 1em plus 0.5em minus 0.4em\relax IEEE, 2022, pp. 1--8.

\bibitem{Ref26}
L.~Lei, T.~Chen, S.~Li, and J.~Li, ``Micro-expression recognition based on facial graph representation learning and facial action unit fusion,'' in \emph{Proceedings of the IEEE/CVF Conference on Computer Vision and Pattern Recognition}, 2021, pp. 1571--1580.

\bibitem{Ref27}
T.~Pu, T.~Chen, Y.~Xie, H.~Wu, and L.~Lin, ``Au-expression knowledge constrained representation learning for facial expression recognition,'' in \emph{2021 IEEE International Conference on Robotics and Automation (ICRA)}.\hskip 1em plus 0.5em minus 0.4em\relax IEEE, 2021, pp. 11\,154--11\,161.

\bibitem{Ref28}
Y.~Liu, X.~Zhang, Y.~Lin, and H.~Wang, ``Facial expression recognition via deep action units graph network based on psychological mechanism,'' \emph{IEEE Transactions on Cognitive and Developmental Systems}, vol.~12, no.~2, pp. 311--322, 2019.

\bibitem{Ref29}
Z.~Cui, T.~Song, Y.~Wang, and Q.~Ji, ``Knowledge augmented deep neural networks for joint facial expression and action unit recognition,'' \emph{Advances in Neural Information Processing Systems}, vol.~33, pp. 14\,338--14\,349, 2020.

\bibitem{Ref36}
L.~Zhou, Q.~Mao, and M.~Dong, ``Objective class-based micro-expression recognition through simultaneous action unit detection and feature aggregation,'' \emph{arXiv preprint arXiv:2012.13148}, 2020.

\bibitem{Ref37}
G.~Peng and S.~Wang, ``Weakly supervised facial action unit recognition through adversarial training,'' in \emph{Proceedings of the IEEE conference on computer vision and pattern recognition}, 2018, pp. 2188--2196.

\bibitem{Ref30}
F.~Ma, B.~Sun, and S.~Li, ``Facial expression recognition with visual transformers and attentional selective fusion,'' \emph{IEEE Transactions on Affective Computing}, 2021.

\bibitem{Ref31}
K.~Wang, X.~Peng, J.~Yang, D.~Meng, and Y.~Qiao, ``Region attention networks for pose and occlusion robust facial expression recognition,'' \emph{IEEE Transactions on Image Processing}, vol.~29, pp. 4057--4069, 2020.

\bibitem{zhang2021joint}
X.~Zhang, F.~Zhang, and C.~Xu, ``Joint expression synthesis and representation learning for facial expression recognition,'' \emph{IEEE Transactions on Circuits and Systems for Video Technology}, vol.~32, no.~3, pp. 1681--1695, 2021.

\bibitem{Ref32}
L.~Wang, X.~Zhang, N.~Jiang, H.~Wu, and J.~Yang, ``D 2 s: Dynamic distribution supervision for multi-label facial expression recognition,'' in \emph{2022 IEEE International Conference on Multimedia and Expo (ICME)}.\hskip 1em plus 0.5em minus 0.4em\relax IEEE, 2022, pp. 1--6.

\bibitem{Ref33}
P.~Ekman and W.~V. Friesen, ``Facial action coding system,'' \emph{Environmental Psychology \& Nonverbal Behavior}, 1978.

\bibitem{Ref38}
W.~Park, D.~Kim, Y.~Lu, and M.~Cho, ``Relational knowledge distillation,'' in \emph{Proceedings of the IEEE/CVF Conference on Computer Vision and Pattern Recognition}, 2019, pp. 3967--3976.

\bibitem{Ref39}
Y.~Zhang, T.~Xiang, T.~M. Hospedales, and H.~Lu, ``Deep mutual learning,'' in \emph{Proceedings of the IEEE conference on computer vision and pattern recognition}, 2018, pp. 4320--4328.

\bibitem{Ref40}
G.~Song and W.~Chai, ``Collaborative learning for deep neural networks,'' \emph{Advances in neural information processing systems}, vol.~31, 2018.

\bibitem{Ref41}
T.~Batra and D.~Parikh, ``Cooperative learning with visual attributes,'' \emph{arXiv preprint arXiv:1705.05512}, 2017.

\bibitem{Ref72}
Z.~Zhao, Q.~Liu, and S.~Wang, ``Learning deep global multi-scale and local attention features for facial expression recognition in the wild,'' \emph{IEEE Transactions on Image Processing}, vol.~30, pp. 6544--6556, 2021.

\bibitem{Ref42}
J.~Duncan, F.~Gosselin, C.~Cobarro, G.~Dugas, C.~Blais, and D.~Fiset, ``Orientations for the successful categorization of facial expressions and their link with facial features,'' \emph{Journal of vision}, vol.~17, no.~14, pp. 7--7, 2017.

\bibitem{Ref43}
F.~Xue, Q.~Wang, and G.~Guo, ``Transfer: Learning relation-aware facial expression representations with transformers,'' in \emph{Proceedings of the IEEE/CVF International Conference on Computer Vision}, 2021, pp. 3601--3610.

\bibitem{Ref44}
D.~Ruan, R.~Mo, Y.~Yan, S.~Chen, J.-H. Xue, and H.~Wang, ``Adaptive deep disturbance-disentangled learning for facial expression recognition,'' \emph{International Journal of Computer Vision}, vol. 130, no.~2, pp. 455--477, 2022.

\bibitem{Ref45}
A.~Kendall and Y.~Gal, ``What uncertainties do we need in bayesian deep learning for computer vision?'' \emph{Advances in neural information processing systems}, vol.~30, 2017.

\bibitem{Ref46}
A.~Dhall, R.~Goecke, S.~Lucey, and T.~Gedeon, ``Collecting large, richly annotated facial-expression databases from movies,'' \emph{IEEE multimedia}, vol.~19, no.~03, pp. 34--41, 2012.

\bibitem{Ref47}
Q.~Cao, L.~Shen, W.~Xie, O.~M. Parkhi, and A.~Zisserman, ``Vggface2: A dataset for recognising faces across pose and age,'' in \emph{2018 13th IEEE international conference on automatic face \& gesture recognition (FG 2018)}.\hskip 1em plus 0.5em minus 0.4em\relax IEEE, 2018, pp. 67--74.

\bibitem{Ref48}
J.~Deng, J.~Guo, E.~Ververas, I.~Kotsia, and S.~Zafeiriou, ``Retinaface: Single-shot multi-level face localisation in the wild,'' in \emph{Proceedings of the IEEE/CVF conference on computer vision and pattern recognition}, 2020, pp. 5203--5212.

\bibitem{Ref51}
H.~Yan, Y.~Gu, X.~Zhang, Y.~Wang, Y.~Ji, and F.~Ren, ``Mitigating label-noise for facial expression recognition in the wild,'' in \emph{2022 IEEE International Conference on Multimedia and Expo (ICME)}.\hskip 1em plus 0.5em minus 0.4em\relax IEEE, 2022, pp. 1--6.

\bibitem{Ref52}
W.~Yu and H.~Xu, ``Co-attentive multi-task convolutional neural network for facial expression recognition,'' \emph{Pattern Recognition}, vol. 123, p. 108401, 2022.

\bibitem{Ref61}
W.~Xie, H.~Wu, Y.~Tian, M.~Bai, and L.~Shen, ``Triplet loss with multistage outlier suppression and class-pair margins for facial expression recognition,'' \emph{IEEE Transactions on Circuits and Systems for Video Technology}, vol.~32, no.~2, pp. 690--703, 2021.

\bibitem{Ref59}
M.~Huang, X.~Zhang, X.~Lan, H.~Wang, and Y.~Tang, ``Convolution by multiplication: Accelerated two-stream fourier domain convolutional neural network for facial expression recognition,'' \emph{IEEE Transactions on Circuits and Systems for Video Technology}, vol.~32, no.~3, pp. 1431--1442, 2021.

\bibitem{Ref60}
Y.~Li, Y.~Lu, M.~Gong, L.~Liu, and L.~Zhao, ``Dual-channel feature disentanglement for identity-invariant facial expression recognition,'' \emph{Information Sciences}, vol. 608, pp. 410--423, 2022.

\bibitem{Ref54}
J.~Shao, Z.~Wu, Y.~Luo, S.~Huang, X.~Pu, and Y.~Ren, ``Self-paced label distribution learning for in-the-wild facial expression recognition,'' in \emph{Proceedings of the 30th ACM International Conference on Multimedia}, 2022, pp. 161--169.

\bibitem{Ref53}
F.~Zhang, M.~Xu, and C.~Xu, ``Weakly-supervised facial expression recognition in the wild with noisy data,'' \emph{IEEE Transactions on Multimedia}, vol.~24, pp. 1800--1814, 2021.

\bibitem{Ref62}
J.~Cai, Z.~Meng, A.~S. Khan, Z.~Li, J.~O'Reilly, and Y.~Tong, ``Probabilistic attribute tree structured convolutional neural networks for facial expression recognition in the wild,'' \emph{IEEE Transactions on Affective Computing}, 2022.

\bibitem{Ref68}
Y.~Zhang, C.~Wang, X.~Ling, and W.~Deng, ``Learn from all: Erasing attention consistency for noisy label facial expression recognition,'' in \emph{European Conference on Computer Vision}.\hskip 1em plus 0.5em minus 0.4em\relax Springer, 2022, pp. 418--434.

\bibitem{Ref64}
D.~Zeng, Z.~Lin, X.~Yan, Y.~Liu, F.~Wang, and B.~Tang, ``Face2exp: Combating data biases for facial expression recognition,'' in \emph{Proceedings of the IEEE/CVF Conference on Computer Vision and Pattern Recognition}, 2022, pp. 20\,291--20\,300.

\bibitem{Ref57}
Y.~Ji, Y.~Hu, Y.~Yang, and H.~T. Shen, ``Region attention enhanced unsupervised cross-domain facial emotion recognition,'' \emph{IEEE Transactions on Knowledge and Data Engineering}, 2021.

\bibitem{Ref56}
A.~H. Farzaneh and X.~Qi, ``Discriminant distribution-agnostic loss for facial expression recognition in the wild,'' in \emph{Proceedings of the IEEE/CVF Conference on Computer Vision and Pattern Recognition Workshops}, 2020, pp. 406--407.

\bibitem{Ref55}
J.~Jiang and W.~Deng, ``Boosting facial expression recognition by a semi-supervised progressive teacher,'' \emph{IEEE Transactions on Affective Computing}, 2021.

\bibitem{Ref63}
F.~Ma, B.~Sun, and S.~Li, ``Robust facial expression recognition with convolutional visual transformers,'' \emph{arXiv preprint arXiv:2103.16854}, 2021.

\end{thebibliography}

\vfill
\end{document}